# Conditional Multidimensional Scaling with Incomplete Conditioning Data

Anh Tuan Bui

Department of Statistical Sciences and Operations Research, Virginia Commonwealth University, 1015 Floyd Avenue, Box 843083, Richmond, Virginia 23284, USA.

Email: buiat2@vcu.edu.

**Abstract:**

Conditional multidimensional scaling seeks for a low-dimensional configuration from pairwise dissimilarities, in the presence of other known features. By taking advantage of available data of the known features, conditional multidimensional scaling improves the estimation quality of the low-dimensional configuration and simplifies knowledge discovery tasks. However, existing conditional multidimensional scaling methods require full data of the known features, which may not be always attainable due to time, cost, and other constraints. This paper proposes a conditional multidimensional scaling method that can learn the low-dimensional configuration when there are missing values in the known features. The method can also impute the missing values, which provides additional insights of the problem. Computer codes of this method are maintained in the **cml** R package on CRAN.

## 1. Introduction

Multidimensional scaling (MDS) is among the most popular dimension reduction methods. Some recent application domains of MDS include manufacturing (Bui and Apley 2022a), geochemistry (Song et al. 2022), sport psychology (Ayala et al. 2022), and earth science (Vermeesch et al. 2023). A salient feature of MDS is that it can learn a low-dimensional configuration for objects from their pairwise dissimilarities. The low-dimensional configuration represents the underlying features that govern the dissimilarities between the objects. On top of being applicable for data that come in dissimilarity form (e.g., similarity ratings or intercorrelations), MDS can be used for any other data formats if meaningful dissimilarity measures can be defined. The flexibility of MDS is especially helpful for data with complex structures, for which the Euclidean space is not directly appropriate. Some examples include distance measures for statistical distributions (Thas 2010), dissimilarity for image data of random heterogeneous materials (Bui and Apley 2021), diversity indices for ecological communities (Oksanen 2024), and dissimilarity between unstructured point clouds (Bui and Apley 2022b), and the cosine dissimilarity commonly used for text data.

In most science and engineering applications, knowledge and data of some underlying features that govern the dissimilarities between the objects are often available or attainable. For instance, Rosenberg and Kim (1975) found that gender, among other features, of 15 common kinship terms explained most of the dissimilarities between the kinship terms. However, the gender feature can be expected (or at least known after this study) to contribute the kinship term dissimilarities. Such underlying features will be referred to as known features in this paper for simplicity.

Bui (2021, 2024) argued that incorporating knowledge of the known features into the dimension reduction process was more advantageous. To begin with, this improves the estimation quality of the low-dimensional configuration by utilizing more fully available data. In addition, this simplifies visualization and knowledge discovery tasks in two aspects. The first aspect is that by marginalizing the known features, we can avoid the known features masking the features of the low-dimensional configuration when visualizing the reduced-dimension space. The second aspect is that the discovered features in the previous analyses can be used as the known features in subsequent analyses. This leads to a more straightforward knowledge discover process because instead of having to recognize all features of the low-dimensional configuration altogether, at least only one feature is needed to be identified in each analysis, repeatedly.

In light of this, *conditional MDS* (Bui 2021, 2024) was proposed to address this limitation. Specifically, Bui (2024) proposed to minimize a *conditional stress* function by *conditional SMACOF*, an iterative optimization algorithm based on majorization (see Section 2.2). Instead of solving the conditional MDS problem via iterative optimization, Bui (2022) proposed a closed-form solution for this problem. This solution is based on the closed-form solutions of multiple linear regression and eigendecomposition. The method of Bui (2022) can be used either as an alternative or for initializing conditional SMACOF (see Appendix A for a brief review).

Nevertheless, these conditional MDS methods require full data of the known features. They are not applicable when the known features have missing values. For example, the value of the gender feature of the cousin term in the aforementioned kinship terms example is undefined. One may exclude objects with missing known feature values so that the existing conditional MDS methods can be used. This is problematic because (*i*) failing to utilize available data generally leads to poorer estimation of the low-dimensional configuration, and (*ii*) excluding objects is unacceptable when their coordinates in the reduced-dimension space are of interest.

To address this limitation of the existing conditional MDS methods, this paper develops a conditional MDS approach that can handle missing values in the known features. The convergence of the proposed method is shown under standard assumptions of the existing conditional MDS methods. By using more data, the new approach improves the learning quality of the low-dimensional configuration in both simulation and real examples. Moreover, it provides sensible dimension reduction results even when the ratio of missing known feature values is large. This implies that practitioners may be able to reduce their costs and efforts by intentionally acquiring fewer data of the known features. The proposed method can also impute the missing known feature values. This may provide more insights into the study problem (e.g., how people perceive the gender of the cousin term in the kinship terms example mentioned above).

The organization of this paper is as follows. Section 2.1 demonstrates the fundamental difference of conditional MDS and related dimension reduction literatures. Section 2.2 summarizes the conditional SMACOF algorithm in Bui (2024) to facilitate understanding of the proposed method. Section 3 presents the proposed method with its theoretical/computational properties. In Section 4, we evaluate the proposed method on a simulated car-brand perception example and the kinship terms example. Section 4 also compares the proposed method with the conditional SMACOF approach of Bui (2024) applied only to the objects that have complete data. Finally, Section 5 concludes this paper.

## 2. Related Work

### 2.1. Dimension Reduction with Additional Information

This section briefly discusses dimension reduction literatures that incorporate additional information to the dimension reduction process. The discussions focus on demonstrating the fundamental difference between conditional MDS and these literatures.

First, the low-dimensional configuration obtained from dimension reduction is widely used as the input of predictive models to predict some response variables. However, the predictive models may be suboptimal if the response variables are not considered in the dimension reduction process (Chao et al. 2019). *Supervised dimension reduction* (see, e.g., Björklund et al. 2023) is a large literature that aims to address this issue. The methods in this literature largely extends popular dimension reduction methods to incorporate the information of the response variables. Chao et al. (2019) categorizes these methods into three main groups based on their dimension reduction methodology: principal component analysis, non-negative matrix factorization, and manifold learning. Methods based on manifold learning techniques that can directly take pairwise dissimilarities as the inputs are more relevant to this paper. Some recent proposals include semi-supervised local multi-manifold ISOMAP (Zhang et al. 2018), supervised t-SNE (Hajderanj et al. 2019; Cheng et al. 2021), and parametric UMAP (Sainburg et al. 2021). See Chao et al. (2019) for a full review of the supervised dimension reduction literature.

Nonetheless, the supervised dimension reduction problem is fundamentally different from the conditional MDS problem. Supervised dimension reduction essentially assumes that a mapping of the low-dimensional configuration to the response variables exists. However, there are no response variables in the conditional MDS problem. The known features in conditional MDS cannot be simply treated as the response variables, because the former does not necessarily depend on the low-dimensional configuration. In the context of conditional MDS, the low-dimensional configuration and the known features altogether govern the observed dissimilarities. This relationship in conditional MDS is not the same as that in supervised dimension reduction. As such, supervised dimension reduction techniques cannot solve the conditional MDS problem.

Another large literature that incorporates the response variables into the dimension reduction procedures is *sufficient dimension reduction*. The methods in this literature (e.g., Zhang et al. 2018; Forzani et al. 2019; Chen et al. 2022) estimate the low-dimensional configuration that is sufficient to predict the response variables. To satisfy the statistical sufficiency, the conditional distribution of the response variable given the learned low dimensional-configuration must be similar to the conditional distribution of the response variable given the original predictor variables. See Cook (2018) for more discussions of the sufficient dimension reduction approaches. Nevertheless, similar to supervised dimension reduction, sufficient dimension reduction assumes that the response variables are functions of the low-dimensional configuration. This is not necessarily the case in the conditional MDS context. Hence, conditional MDS completely differs from sufficient dimension reduction, and the latter is not the solution of the conditional MDS problems.

A more related methodology to conditional MDS is *interactive dimension reduction* (see, e.g., Fujiwara et al. (2021), Bian et al. (2021), and the references therein). Interactive dimension reduction also integrates domain knowledge to the dimension reduction procedure. However, interactive dimension reduction actively involves practitioners' interaction during the dimension reduction procedure (such as changing features weights, model parameters, or outputs) to produce comprehensible dimension reduction results. This is different from conditional MDS, which does not require such human interaction during the dimension reduction process.

## 2.2. Conditional MDS via Conditional SMACOF

This section briefly reviews the conditional MDS method based on conditional SMACOF in Bui (2024). Let $\{\mathbf{u}_i \in \mathbb{R}^p: i = 1, 2, ..., N\}$ and $\{\mathbf{v}_i \in \mathbb{R}^q: i = 1, 2, ..., N\}$ be the values of $p$ features of the low-dimensional configuration and $q$ known features, respectively, of $N$ objects. The goal of conditional MDS is finding a low-dimensional configuration $\mathbf{U} = [\mathbf{u}_1, \mathbf{u}_2, ..., \mathbf{u}_N]^T \in \mathbb{R}^{N \times p}$ from given dissimilarities $\delta_{ij}$ $(i, j = 1, ..., N)$ and data of the known features $\mathbf{V} = [\mathbf{v}_1, \mathbf{v}_2, ..., \mathbf{v}_N]^T \in \mathbb{R}^{N \times q}$.

To ensure that the low-dimensional configuration $\mathbf{U}$ and the known features $\mathbf{V}$ are compatible in the same coordinate system of the reduced-dimension space, consider $\tilde{\mathbf{v}}_i = \mathbf{B}^T \mathbf{v}_i$ ($i =$ $1, 2, \ldots, N$), where $\mathbf{B} \in \mathbb{R}^{q \times q}$. Note that $\mathbf{B}$ represents the joint effects of the known features on the pairwise distances in the reduced-dimension space, let denoted by $d_{ij}(\mathbf{U}, \widetilde{\mathbf{V}}) = \sqrt{\|\mathbf{u}_i - \mathbf{u}_j\|^2 + \|\tilde{\mathbf{v}}_i - \tilde{\mathbf{v}}_j\|^2}$, where $\widetilde{\mathbf{V}} = [\tilde{\mathbf{v}}_1, \tilde{\mathbf{v}}_2, \ldots, \tilde{\mathbf{v}}_N]^T = \mathbf{VB}$. It is worthy to emphasize that the Euclidean distance is used for the reduced-dimension space here, as with metric MDS and many other dimension reduction methods. This will be the basis for relevant theoretical results in this paper. These results do not necessarily hold for non-Euclidean measures.

To learn $\mathbf{U}$ and $\mathbf{B}$ jointly, Bui (2024) proposed to minimize the *conditional stress function*

$$\sigma(\mathbf{U}, \widetilde{\mathbf{V}}) = \sum_{i<j} w_{ij} \left( \tilde{\delta}_{ij} - d_{ij}(\mathbf{U}, \widetilde{\mathbf{V}}) \right)^2 \tag{1}$$

over $\mathbf{U}$ and $\mathbf{B}$, where the $w_{ij}$'s are given weights, the $\tilde{\delta}_{ij}$'s are monotonic transformations of the given $\delta_{ij}$'s. Bui (2024) developed an algorithm called *conditional SMACOF* to solve this problem. This algorithm is based on SMACOF (de Leeuw 1977), a majorization-based optimization technique commonly used in metric MDS.

To understand conditional SMACOF, denote $\mathbf{H} = [h_{ij}]_{i,j=1}^N \in \mathbb{R}^{N \times N}$ with $h_{ij} = -w_{ij}$ if $i \neq j$ and $h_{ii} = \sum_{j=1, j \neq i}^N w_{ij}$. Let $\mathbf{H}^+$ be the Moore-Penrose inverse of $\mathbf{H}$. As noted on page 191 of Borg and Groenen (2005), we can compute $\mathbf{H}^+$ via matrix inversion: $\mathbf{H}^+ = (\mathbf{H} + \mathbf{1}_{N \times N})^{-1} - N^{-2} \mathbf{1}_{N \times N}$, where $\mathbf{1}_{N \times N}$ is an $N \times N$ matrix with all elements equal to 1. Additionally, denote $\mathbf{C} = [c_{ij}]_{i,j=1}^N$ by

$$c_{ij} = \begin{cases} -\frac{w_{ij} \tilde{\delta}_{ij}}{d_{ij}(\mathbf{U}, \widetilde{\mathbf{V}})} & \text{if } d_{ij}(\mathbf{U}, \widetilde{\mathbf{V}}) \neq 0 \\ 0 & \text{if } d_{ij}(\mathbf{U}, \widetilde{\mathbf{V}}) = 0 \end{cases} \text{ for } i \neq j,$$

and $c_{ii} = \sum_{j=1, j \neq i}^N c_{ij}$. Conditional SMACOF iteratively updates $\mathbf{U}$ and $\mathbf{B}$ by

$$\mathbf{U}^{[l]} = \mathbf{H}^+ \mathbf{C}^{[l-1]} \mathbf{U}^{[l-1]} \text{ and } \mathbf{B}^{[l]} = (\mathbf{V}^T \mathbf{H} \mathbf{V})^{-1} \mathbf{V}^T \mathbf{C}^{[l-1]} \mathbf{V} \mathbf{B}^{[l-1]},$$

where the superscript [*l*] denotes the value of the corresponding quantity at the *l*th iteration. Bui (2024) showed that conditional SMACOF converges to a critical point of the conditional stress function under Assumption 1.

***Assumption 1***:

a) *The dissimilarities and the weights $w_{ij}$ (i, j = 1, 2, ..., N) are symmetric.*
b) *The weight matrix $[w]_{i,j=1}^{N}$ does not have groups of objects for which the intergroup weights are always 0, i.e., it is irreducible.*
c) ***V*** *contains q linearly independent difference vectors $\boldsymbol{v}_j - \boldsymbol{v}_k$ ($1 \leq j < k \leq N$).*

It is relatively straightforward to satisfy Assumption 1. For Assumption 1(a), we can symmetrize the $\delta_{ij}$'s by replacing each $\delta_{ij}$ and $\delta_{ji}$ with $\delta_{ij} + \delta_{ji}$. And symmetric $w_{ij}$'s are generally used for symmetric dissimilarities. For Assumption 1(b), if the weight matrix is reducible, it has groups of objects with intergroup weights of 0, and it can be reduced to irreducible matrices. In this case, we can apply conditional SMACOF to each resulting irreducible matrix. For Assumption 1(c), a violation is practically rare because the number of the difference vectors increases quadratically with *N*. If needed, we can satisfy this assumption by either reducing *q* or perturbing **V** slightly to contain *q* linearly independent difference vectors.

Because practically good solutions can be obtained way before the convergence, Bui (2024) recommended to stop the iterative updates when the reduction in the conditional stress is small. In this case, the *normalized conditional stress*

$$\sigma_n(\mathbf{U}, \widetilde{\mathbf{V}}) = \frac{\sigma(\mathbf{U}, \widetilde{\mathbf{V}})}{\sum_{i<j} w_{ij} \widetilde{\delta}_{ij}^2} = \frac{\sum_{i<j} w_{ij} \left(\widetilde{\delta}_{ij} - d_{ij}(\mathbf{U}, \widetilde{\mathbf{V}})\right)^2}{\sum_{i<j} w_{ij} \widetilde{\delta}_{ij}^2}, \qquad (2)$$

is often used instead of the conditional stress. Because the former does not depend on the scale of the $\widetilde{\delta}_{ij}$'s as the latter does, one can use a common threshold for the reduction in the normalized conditional stress in different problems.

As with the metric MDS framework, it is possible to extend conditional MDS to nonlinear dimension reduction via two approaches. The first approach uses a local weighting scheme for the weights $w_{ij}$'s to focus on the local behavior of the dissimilarities. For example, Sammon (1969) proposed $w_{ij} = 1/\left(\tilde{\delta}_{ij} \sum_{i<j} \tilde{\delta}_{ij}\right) \forall i, j$, and Demartines and Herault (1997) set each $w_{ij}$ as some decreasing function of the distance between the *i*th and *j*th objects in the reduced-dimension space. The second approach involves transformations of the dissimilarities such that linear methods become applicable for nonlinear dimension reduction. A seminal example of such a transformation is the geodesic distance transformation used in ISOMAP (Tenenbaum et al. 2000). An alternative approach is kernel-based transformation such as those in Ding and Ma (2023) and Ding and Wu (2023). It should be noted that these transformations were originally proposed for Euclidean distances. A simple solution to satisfy this condition is to use MDS to extract vector data (not necessarily low-dimensional) from the given dissimilarities so that Euclidean distances can be calculated from them. Nevertheless, future studies to extend those transformations directly to dissimilarities would be of interest.

However, the conditional MDS approaches in Bui (2022, 2024) require complete data of each $\mathbf{v}_i$ (*i* = 1, 2, ..., *N*), as discussed in Section 1. They are not applicable when there are missing known feature values. This calls for a new conditional MDS approach to address this problem. The following section presents the development of such a method.

## 3. Conditional MDS with Missing Known Feature Values

This section introduces the proposed method for handling missing known feature values for conditional MDS. We first present the proposed method for general cases when the weights $w_{ij}$'s are arbitrary. These cases often arise in two scenarios. First, the weights $w_{ij}$'s corresponding to the missing dissimilarities $\delta_{ij}$'s are 0 (whereas the others are positive). Second, when a local weighting scheme (e.g., as in Sammon mapping by Sammon (1969)) is employed for non-linear dimension reduction.

Without loss of generality, assume that the observation indices are sorted such that $\mathbf{V} = [\mathbf{V}_1^T, \mathbf{V}_2^T]^T$, where $\mathbf{V}_1 \in \mathbb{R}^{N_1 \times q}$ ($N_1 < N$) contains all the complete observations, and $\mathbf{V}_2$ contains the other $N_2 = N - N_1$ cases with missing data. Furthermore, let $\widetilde{\mathbf{V}}_2$ be such that $\widetilde{\mathbf{V}} = [\mathbf{B}^T\mathbf{V}_1^T, \widetilde{\mathbf{V}}_2^T]^T$. The proposed method aims to minimize the conditional stress in Bui (2024) over $\mathbf{U}$, $\mathbf{B}$, and $\widetilde{\mathbf{V}}_2$:

$$\min_{\mathbf{U},\mathbf{B},\widetilde{\mathbf{V}}_2} \sum_{i<j} w_{ij} \left( \tilde{\delta}_{ij} - d_{ij}\left(\mathbf{U}, [\mathbf{B}^T\mathbf{V}_1^T, \widetilde{\mathbf{V}}_2^T]^T\right) \right)^2. \tag{3}$$

To find the solutions of $\mathbf{U}$, $\mathbf{B}$, and $\widetilde{\mathbf{V}}_2$, the proposed method iteratively updates these variables based on the formulas in (4), (5), and (6) in Theorem 1, respectively. Theorem 1 implies the convergence of this iterative updating strategy because the resulting generated sequence of conditional stress values is strictly decreasing until no further reduction can be attained.

**Theorem 1**: *Decompose $\mathbf{H}$ into $\mathbf{H} = \begin{bmatrix} \mathbf{H}_{11} & \mathbf{H}_{12} \\ \mathbf{H}_{21} & \mathbf{H}_{22} \end{bmatrix}$, and similarly, $\mathbf{C} = \begin{bmatrix} \mathbf{C}_{11} & \mathbf{C}_{12} \\ \mathbf{C}_{21} & \mathbf{C}_{22} \end{bmatrix}$, where $\mathbf{H}_{11}$ and $\mathbf{C}_{11}$ are of size $N_1 \times N_1$. Let $\mathbf{K}_b = \mathbf{S} - \mathbf{V}_1^T \mathbf{H}_{12} \mathbf{H}_{22}^{-1} \mathbf{H}_{21} \mathbf{V}_1$ and $\mathbf{K}_{\tilde{v}_2} = \mathbf{H}_{22} - \mathbf{G}\mathbf{V}_1^T \mathbf{H}_{12}$, where $\mathbf{S} = \mathbf{V}_1^T \mathbf{H}_{11} \mathbf{V}_1$ and $\mathbf{G} = \mathbf{H}_{21} \mathbf{V}_1 \mathbf{S}^{-1}$. If $\mathbf{U}^{[l]}$, $\mathbf{B}^{[l]}$, and $\widetilde{\mathbf{V}}_2^{[l]}$ are updated by*

$$\mathbf{U}^{[l]} = \mathbf{H}^+ \mathbf{C}^{[l-1]} \mathbf{U}^{[l-1]} \tag{4}$$

$$\mathbf{B}^{[l]} = \mathbf{K}_b^{-1} \mathbf{V}_1^T \left[ \left(\mathbf{C}_{11}^{[l-1]} - \mathbf{H}_{12}\mathbf{H}_{22}^{-1}\mathbf{C}_{21}^{[l-1]}\right)\mathbf{V}_1 \mathbf{B}^{[l-1]} + \left([\mathbf{C}_{12}^{[l-1]} - \mathbf{H}_{12}\mathbf{H}_{22}^{-1}\mathbf{C}_{22}^{[l-1]}\right)\widetilde{\mathbf{V}}_2^{[l-1]} \right] \tag{5}$$

$$\widetilde{\mathbf{V}}_2^{[l]} = \mathbf{K}_{\tilde{v}_2}^{-1} \left[ \left(\mathbf{C}_{21}^{[l-1]} - \mathbf{G}\mathbf{V}_1^T\mathbf{C}_{11}^{[l-1]}\right)\mathbf{V}_1 \mathbf{B}^{[l-1]} + \left(\mathbf{C}_{22}^{[l-1]} - \mathbf{G}\mathbf{V}_1^T\mathbf{C}_{12}^{[l-1]}\right)\widetilde{\mathbf{V}}_2^{[l-1]} \right], \tag{6}$$

*under Assumption 1,* $\sigma\left(\mathbf{U}^{[l]}, \left[\left(\mathbf{B}^{[l]}\right)^T\mathbf{V}_1^T, \left(\widetilde{\mathbf{V}}_2^{[l]}\right)^T\right]^T\right) \leq \sigma\left(\mathbf{U}^{[l-1]}, \left[\left(\mathbf{B}^{[l-1]}\right)^T\mathbf{V}_1^T, \left(\widetilde{\mathbf{V}}_2^{[l-1]}\right)^T\right]^T\right)$, *and the equality occurs when* $\mathbf{U}^{[l]} = \mathbf{U}^{[l-1]}$, $\mathbf{B}^{[l]} = \mathbf{B}^{[l-1]}$, *and* $\widetilde{\mathbf{V}}_2^{[l]} = \widetilde{\mathbf{V}}_2^{[l-1]}$.

The proof of Theorem 1 (see Appendix D) provides insights on how the update formulas (4—6) were derived. The underlying idea is to minimize the non-convex conditional stress function via minimizing a series of majorizing functions (as functions of $\mathbf{U}$, $\mathbf{B}$, and $\widetilde{\mathbf{V}}_2$ and their current values) of the conditional stress. The majorizing functions are convex and quadratic in $\mathbf{U}$, $\mathbf{B}$, and $\widetilde{\mathbf{V}}_2$. Therefore, their global minima of $\mathbf{U}$, $\mathbf{B}$, and $\widetilde{\mathbf{V}}_2$ can be found via setting their derivatives to 0. Note that the proof of Theorem 1 is based on Assumption 1 (the assumptions of conditional SMACOF) and Lemmas 1—2 (see Appendices B—C). The proofs of these lemmas can also be found in Appendices B—C.

Algorithm 1 summarizes the main steps of the proposed method. Step 1 of Algorithm 1 first initializes $\mathbf{U}^{[0]}$, $\mathbf{B}^{[0]}$, and $\widetilde{\mathbf{V}}_2^{[0]}$. A naive initialization method is setting $\mathbf{B}^{[0]}$ as an identity matrix and using random values for $\mathbf{U}^{[0]}$ and $\widetilde{\mathbf{V}}_2^{[0]}$. Alternatively, let $\mathbf{U}_1$ and $\mathbf{U}_2$ be the blocks in $\mathbf{U}$ that correspond to the $\mathbf{V}_1$ and $\mathbf{V}_2$ blocks in $\mathbf{V}$, respectively. We may also initialize $\mathbf{B}^{[0]}$ and $\mathbf{U}_1^{[0]}$ via applying the conditional SMACOF algorithm in Bui (2024) or the closed-form solution in Bui (2022) to the complete data. Then, $\mathbf{U}_1^{[0]}$ can be initialized arbitrarily, and the missing values in $\widetilde{\mathbf{V}}_2^{[0]}$ can be set to 0. Next, Step 1 pre-computes $\mathbf{H}^+$, $\mathbf{H}_{12}\mathbf{H}_{22}^{-1}$, $\mathbf{K}_b^{-1}\mathbf{V}_1^T$, $\mathbf{G}\mathbf{V}_1^T$, and $\mathbf{K}_{\widetilde{v}_2}^{-1}$, which will be used repeatedly in the iterative updates (4—6) in Step 2. Finally, Step 1 calculates the initial normalized conditional stress $\sigma_n^{[0]}$ and sets $\sigma_n^{[-1]} = \infty$. Step 2 of Algorithm 1 updates $\mathbf{U}^{[l]}$, $\mathbf{B}^{[l]}$, and $\widetilde{\mathbf{V}}_2^{[l]}$ by (4), (5), and (6), respectively, for $l = 1, 2, \ldots$, until either it reaches the pre-set maximum number of iterations $l_{max}$ or the reduction in the normalized conditional stress between two consecutive iterations is not greater than the pre-defined threshold $\gamma$.

**Algorithm 1: Arbitrary Weights**

**Inputs:** $\tilde{\Delta}$, $\mathbf{V}_1$, $w_{ij}$ ($\forall i, j$), $l_{max}$, and $\gamma$

**Step 1:**

a) Initialize $\mathbf{U}^{[0]}$, $\mathbf{B}^{[0]}$, and $\tilde{\mathbf{V}}_2^{[0]}$

b) Pre-compute $\mathbf{H}^+$, $\mathbf{H}_{12}\mathbf{H}_{22}^{-1}$, $\mathbf{K}_b^{-1}\mathbf{V}_1^T$, $\mathbf{GV}_1^T$, and $\mathbf{K}_{\tilde{v}_2}^{-1}$

c) Calculate the initial normalized conditional stress $\sigma_n^{[0]}$ by (3)

d) Set $\sigma_n^{[-1]} = \infty$ and $l = 0$

**Step 2:** While $l < l_{max}$ and $\sigma_n^{[l-1]} - \sigma_n^{[l]} > \gamma$ do:

a) $l \leftarrow l + 1$

b) Update $\mathbf{U}^{[l]}$, $\mathbf{B}^{[l]}$, and $\tilde{\mathbf{V}}_2^{[l]}$ by (4), (5), and (6), respectively

c) Calculate the current normalized conditional stress $\sigma_n^{[l]}$ by (3)

**Outputs:** $\mathbf{U}^* = \mathbf{U}^{[l]}$, $\mathbf{B}^* = \mathbf{B}^{[l]}$, $\tilde{\mathbf{V}}_2^* = \tilde{\mathbf{V}}_2^{[l]}$, $\sigma_n^* = \sigma_n^{[l]}$

**Remark 2:** If $\mathbf{B}$ is invertible, we can impute $\mathbf{V}_2$ by $\hat{\mathbf{V}}_2 = \tilde{\mathbf{V}}_2 \mathbf{B}^{-1}$. Note that this formula is based on the premise of this paper that an object has missing values in all known features. When an object has missing values not in all known features, we can use the following corrected formula:

$$\hat{\mathbf{V}}_2 = \mathbf{V}_2 \circ (\mathbf{1}_{N_2 \times q} - \mathbf{M}) + \left[\left(\tilde{\mathbf{V}}_2 - \left(\mathbf{V}_2 \circ (\mathbf{1}_{N_2 \times q} - \mathbf{M})\right)\mathbf{B}\right) \circ \mathbf{M}\right]\mathbf{B}^{-1}, \qquad (7)$$

where $\circ$ is the Hadamard product and $\mathbf{M} = [m_{ij}]_{i=1..N_2; j=1..q}$, with $m_{ij} = 1$ if $[\mathbf{V}_2]_{ij}$ is missing and $m_{ij} = 0$ otherwise for $\forall i, j$. This formula ensures that $[\hat{\mathbf{V}}_2]_{ij} = [\mathbf{V}_2]_{ij}$ if $[\mathbf{V}_2]_{ij}$ is not missing, and the information of $[\mathbf{V}_2]_{ij}$ is used when imputing other missing values. And it reduces to $\hat{\mathbf{V}}_2 = \tilde{\mathbf{V}}_2 \mathbf{B}^{-1}$ if missing values occur in all known features.

**Remark 3:** A common weighting scheme in MDS applications is that all the weights $w_{ij}$'s are equal. This can be viewed as a special case of the arbitrary weighting scheme discussed above. Without loss of generality, we assume that all the weights $w_{ij}$'s are 1. In this case, we can simplify many quantities in the update formulas (4—6) as in Theorem 2 (see Appendix E) to reduce the computational time expenses. Appendix E also provides the proof for Theorem 2. Theorem 2 implies that we will only need to perform a single inversion operation for the $q \times q$ matrix $\mathbf{V}_1^T \mathbf{V}_1$. We can also simplify or avoid some matrix multiplication operations by taking advantage of the special natures of the identity and $\mathbf{1}$ matrices. As a result, the update formulas (4—6) can be simplified as in Corollary 1, which follows immediately after plugging the results in Theorem 2 into Theorem 1. Algorithm 2 summarizes the main steps of the proposed method in the equal weight cases, using the simplifications in Corollary 1.

**Corollary 1**: *Suppose that all the weights $w_{ij}$'s are 1. Let $\mathbf{U}_a^{[l]} = \frac{1}{N} \mathbf{C}^{[l-1]} \mathbf{U}^{[l-1]}$ and $\widetilde{\mathbf{V}}_{2,a}^{[l]} =$*

$\left[ \mathbf{C}_{21}^{[l-1]} + \mathbf{1}_{N_2 \times 1} \left( [\mathbf{v}_1^s]^T \mathbf{S}^{-1} \mathbf{V}_1^T \mathbf{C}_{11}^{[l-1]} \right) \right] (\mathbf{V}_1 \mathbf{B}^{[l-1]}) + \left[ \mathbf{C}_{22}^{[l-1]} + \mathbf{1}_{N_2 \times 1} \left( [\mathbf{v}_1^s]^T \mathbf{S}^{-1} \mathbf{V}_1^T \mathbf{C}_{12}^{[l-1]} \right) \right] \widetilde{\mathbf{V}}_2^{[l-1]}$.

*Additionally, denote $\mathbf{u}_a^{s,[l]}$, $\mathbf{c}_{21}^{s,[l-1]}$ and $\mathbf{c}_{22}^{s,[l-1]}$, and $\widetilde{\mathbf{v}}_{2,a}^{s,[l]}$ as the vectors of the column sums of $\mathbf{U}_a^{[l]}$, $\mathbf{C}_{21}^{[l-1]}$, $\mathbf{C}_{22}^{[l-1]}$, and $\widetilde{\mathbf{V}}_{2,a}^{[l]}$, respectively. Furthermore, $g = 1 + [\mathbf{v}_1^s]^T \mathbf{S}^{-1} \mathbf{v}_1^s$ as defined in Theorem 2. If $\mathbf{U}^{[l]}$, $\mathbf{B}^{[l]}$, and $\widetilde{\mathbf{V}}_2^{[l]}$ are updated by*

$$\mathbf{U}^{[l]} = \mathbf{U}_a^{[l]} - \frac{1}{N} \mathbf{1}_{N \times 1} \left[ \mathbf{u}_a^{s,[l]} \right]^T, \tag{8}$$

$$\mathbf{B}^{[l]} = \mathbf{K}_b^{-1} \mathbf{V}_1^T \left[ \left( \mathbf{C}_{11}^{[l-1]} + \frac{\mathbf{1}_{N_1 \times 1} \left[ \mathbf{c}_{21}^{s,[l-1]} \right]^T}{N_1} \right) \mathbf{V}_1 \mathbf{B}^{[l-1]} + \left( \mathbf{C}_{12}^{[l-1]} + \frac{\mathbf{1}_{N_1 \times 1} \left[ \mathbf{c}_{22}^{s,[l-1]} \right]^T}{N_1} \right) \widetilde{\mathbf{V}}_2^{[l-1]} \right], \tag{9}$$

$$\widetilde{\mathbf{V}}_2^{[l]} = \frac{1}{N} \widetilde{\mathbf{V}}_{2,a}^{[l]} + \frac{g}{N - g N_2} \mathbf{1}_{N_2 \times 1} \left[ \widetilde{\mathbf{v}}_{2,a}^{s,[l]} \right]^T, \tag{10}$$

*then under Assumption 1, $\sigma(\mathbf{U}^{[l]}, \widetilde{\mathbf{V}}^{[l]}) \leq \sigma(\mathbf{U}^{[l-1]}, \widetilde{\mathbf{V}}^{[l-1]})$, and the equality occurs when $\mathbf{U}^{[l]} = \mathbf{U}^{[l-1]}$, $\mathbf{B}^{[l]} = \mathbf{B}^{[l-1]}$, and $\widetilde{\mathbf{V}}_2^{[l]} = \widetilde{\mathbf{V}}_2^{[l-1]}$.*

**Algorithm 2: Equal Weights (Special Case of Algorithm 1)**

**Inputs:** $\widetilde{\Delta}$, $\mathbf{V}_1$, $l_{max}$, and $\gamma$

**Step 1:**

   a) Initialize $\mathbf{U}^{[0]}$, $\mathbf{B}^{[0]}$, and $\widetilde{\mathbf{V}}_2^{[0]}$

   b) Pre-compute $[\mathbf{v}_1^s]^T \mathbf{S}^{-1} \mathbf{V}_1^T$, $\mathbf{K}_b^{-1} \mathbf{V}_1^T$, and $g$

   c) Calculate the initial normalized conditional stress $\sigma_n^{[0]}$ by (2)

   d) Set $\sigma_n^{[-1]} = \infty$ and $l = 0$

**Step 2:** While $l < l_{max}$ and $\sigma_n^{[l-1]} - \sigma_n^{[l]} > \gamma$ do:

   d) $l \leftarrow l + 1$

   e) Update $\mathbf{U}^{[l]}$, $\mathbf{B}^{[l]}$, and $\widetilde{\mathbf{V}}_2^{[l]}$ by (8), (9), and (10), respectively

   f) Calculate the current normalized conditional stress $\sigma_n^{[l]}$ by (2)

**Outputs:** $\mathbf{U}^* = \mathbf{U}^{[l]}$, $\mathbf{B}^* = \mathbf{B}^{[l]}$, $\widetilde{\mathbf{V}}_2^* = \widetilde{\mathbf{V}}_2^{[l]}$, $\sigma_n^* = \sigma_n^{[l]}$

---

To show more clearly the computational benefit of the update formulas in Corollary 1 for the equal weight cases, Table 1 compares the time complexities of the main calculations of the proposed method for arbitrary weights (Algorithm 1) and equal weights (Algorithm 2). Note that the exponents of 2.372 in Table 1 is from the time complexity of the matrix inversion operation using the fastest algorithms (e.g., see Williams et al. 2024). For simplicity, suppose that $p$ and $q$ do not increase with $N$. It can be seen from Table 1 that the time complexity of Algorithm 1 is $O(N^{2.372} + NN_1N_2)$. When $N_1$ or $N_2$ is small and does not increase with $N$, the time complexity of Algorithm 1 reduces to $O(N^{2.372})$. However, when $N_1 = N_2 = N/2$, the time complexity of Algorithm 1 is $O(N^3)$. In contrast, the time complexity of Algorithm 2 is always $O(N^2)$.

**Table 1.** Time complexities of the main calculations of the proposed method in both general and equal weighting schemes.

|  | Calculation | Arbitrary weights | Equal weights |
|---|---|---|---|
| Step 1 | $\mathbf{H}^+$ | $O(N^{2.372})$ | *unnecessary* |
|  | $\mathbf{H}_{21}^T \mathbf{H}_{22}^{-1}$ | $O(N_2^{2.372} + N_1 N_2^2)$ |  |
|  | $\mathbf{G} \mathbf{V}_1^T$ | $O(q^2[NN_2 + q^{0.373}])$ |  |
|  | $\mathbf{K}_{\tilde{v}_2}^{-1}$ | $O(NN_2 + N_2^{2.373})$ |  |
|  | $\mathbf{K}_b^{-1} \mathbf{V}_1^T$ | $O(qN_1^2 + 2qN_1N_2 + q^{2.372})$ | $O(q^2 N_1 + q^{2.372})$ |
|  | $\sigma_n^{[0]}(\mathbf{U}, \widetilde{\mathbf{V}})$ | $O(N^2)$ | $O(N^2)$ |
| Each iteration in Step 2 | $\mathbf{C}$ | $O(N^2[p+q])$ | $O(N^2[p+q])$ |
|  | $\mathbf{U}$ | $O(N^2 p)$ | $O(N^2 p)$ |
|  | $\mathbf{B}$ | $O([N_2 N + qN + q^2]N_1)$ | $O(N^2 + q^2 N_1)$ |
|  | $\widetilde{\mathbf{V}}_2$ | $O([N_1 N + qN + qN_2]N_2)$ | $O(N^2)$ |
|  | $\sigma_n(\mathbf{U}, \widetilde{\mathbf{V}})$ | $O(N^2)$ | $O(N^2)$ |

**Remark 4:** If the column averages of $\mathbf{V}_1$ are 0, we have that $\mathbf{K}_b^{-1} = \frac{1}{N}(\mathbf{V}_1^T \mathbf{V}_1)^{-1}$, $\widetilde{\mathbf{V}}_{2,a}^{[l]} = \mathbf{C}_{21}^{[l-1]} \mathbf{V}_1 \mathbf{B}^{[l-1]} + \mathbf{C}_{22}^{[l-1]} \widetilde{\mathbf{V}}_2^{[l-1]}$, and $g = 1$. Thus, we can center $\mathbf{V}_1$ to simplify further the computations for $\mathbf{K}_b^{-1}$ and $\widetilde{\mathbf{V}}_2^{[l]}$. Note that while centering $\mathbf{V}_1$ will not theoretically change the solution, the actual solution will be likely different due to machine precision, the effect of which would be amplified iteration after iteration. Additionally, we can simplify the update formula (8) by approximating $\boldsymbol{U}^{[l]} \approx \boldsymbol{U}_a^{[l]}$ when $N$ is large. Nevertheless, these simplifications do not alter the $O(N^2)$ time complexity of Algorithm 2.

## 4. Examples

### 1.1. Car-Brand Perception Simulation Example

This section conducts a Monte Carlo study to evaluate the proposed method on the car-brand perception simulation example in Bui (2024). In each replicate, pairwise dissimilarities for $N = 100$ car brands are generated using the weighted Euclidean distances of seven features: Quality, Safety, Value, Performance, Eco, Design, and Technology. The weights of these features are 90/562, 88/562, 83/562, 82/562, 81/562, 70/562, and 68/562, respectively. The numerators of these weights are from the 2014 Car-Brand Perception Survey of Consumer Reports (Bui 2024), and the denominators are such that the weights sum up to 1.

In this example, the values of the features for the car brands are drawn from a Uniform(0, 1) distribution. The weighted Euclidean distances are then added with zero-mean Gaussian noises. The standard deviations of the noises are equal to $\xi_1 = 20\%$ of the corresponding weighted Euclidean distances. We also add to the feature values with zero-mean Gaussian noises (after calculating the weighted Euclidean distances). The standard deviations of these noises are equal to $\xi_2 = 5\%$ of the corresponding feature values. The first four features (Quality, Safety, Value, and Performance) will be treated as known, and the other three features (Eco, Design, and Technology) will be treated as unknown.

We consider four different performance measures. The first and second are the average canonical correlation (ACC) and the Procrustes statistic (PS) between the learned $\mathbf{U}$ and the generated values, respectively. An ACC value close to 1 or a PS value close to 0 suggest that the learned $\mathbf{U}$ is close to the ground truth. The third and the fourth are the mean squared errors of the learned $\mathbf{B}$ (MSE$_B$) and $\mathbf{V}_2$ (MSE$_V$) from their ground truths, respectively.

We compare these performance measures of the proposed method with those of conditional SMACOF (Bui 2024) applied to complete data, using different $N_1/N$ ratios. The different ratios are attained by treating the known feature values of $(N - N_1)$ random generated observations as missing. We consider initializing the proposed method by the naïve approach and applying conditional SMACOF to complete observations (as discussed in Section 3). Figure 1 plots the medians of the four performance measures over 100 Monte Carlo replicates of these methods.

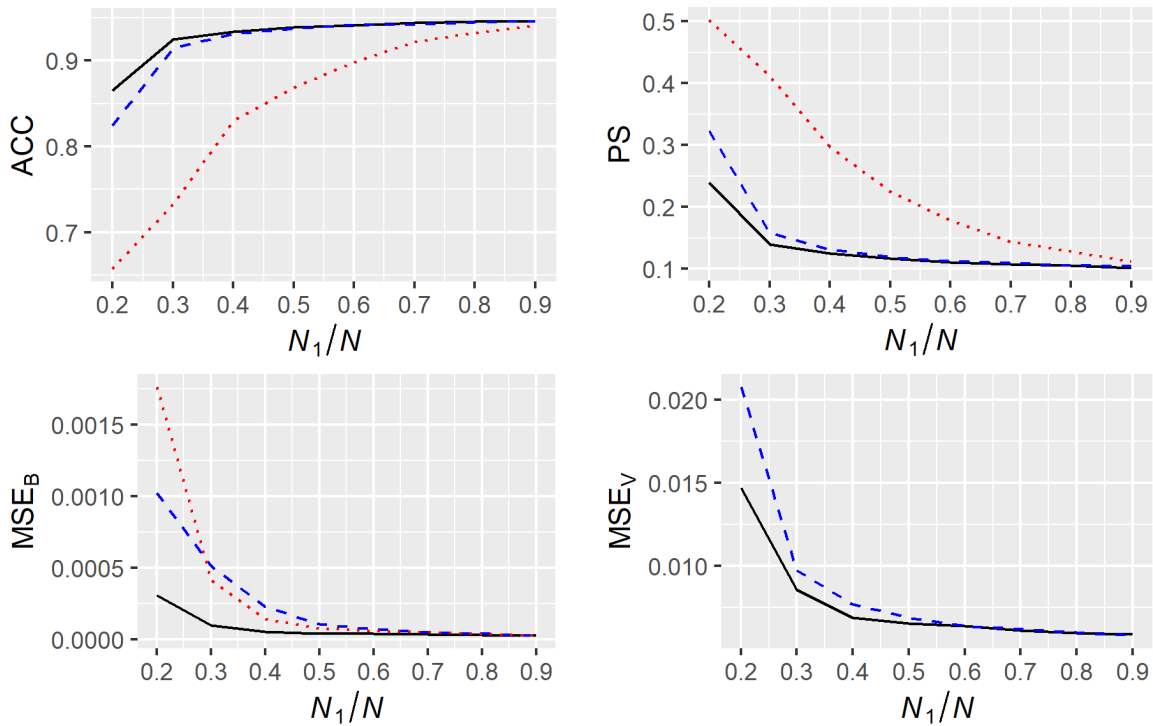

**Figure 1.** Plotting median performance over 100 Monte Carlo replicates against $N_1/N$ in the car-brand perception example of: conditional MDS applied to complete data (dotted curves), the proposed method with naïve initialization (dashed curves), and the proposed method initialized by applying conditional MDS to the complete data (solid curves), for $N = 100$, $\xi_1 = 20\%$, and $\xi_2 = 5\%$.

We can make several observations from Figure 1. First, the proposed method is better than the conditional SMACOF approach applied to complete data (dotted curves) by large margins, using either the naive initialization (dashed curves) or conditional SMACOF applied to complete data for initialization (solid curves). This result is not surprising since the proposed method utilizes more fully available data than the latter approach. Note that the latter approach cannot impute $\mathbf{V}_2$, and thus, it does not have the $MSE_V$ performance (in the bottom right panel of Figure 1). Additionally, the proposed method maintains quite well the learning quality even when $N_1/N$ is low, i.e., the missing data ratio is high. This implies that practitioners can reduce costs and efforts in obtaining data for the known features, by only acquiring data of the known features for a subgroup of objects (the car-brands in this example). Finally, it appears that using conditional SMACOF applied to complete data for initialization of the proposed method is especially helpful when $N_1/N$ is low.

To test the robustness of the methods with noise, we redo the experiment above with double noise levels: $\xi_1 = 40\%$ and $\xi_2 = 10\%$. Figure 2 shows the results of this experiments in the same manner with Figure 1. We can see from Figure 2 that the performances of all methods degrade with this high level of noises. However, the proposed method is still better than conditional SMACOF (Bui 2024) applied to the complete data. The difference between the two initialization methods for the proposed approach is unclear in this case though.

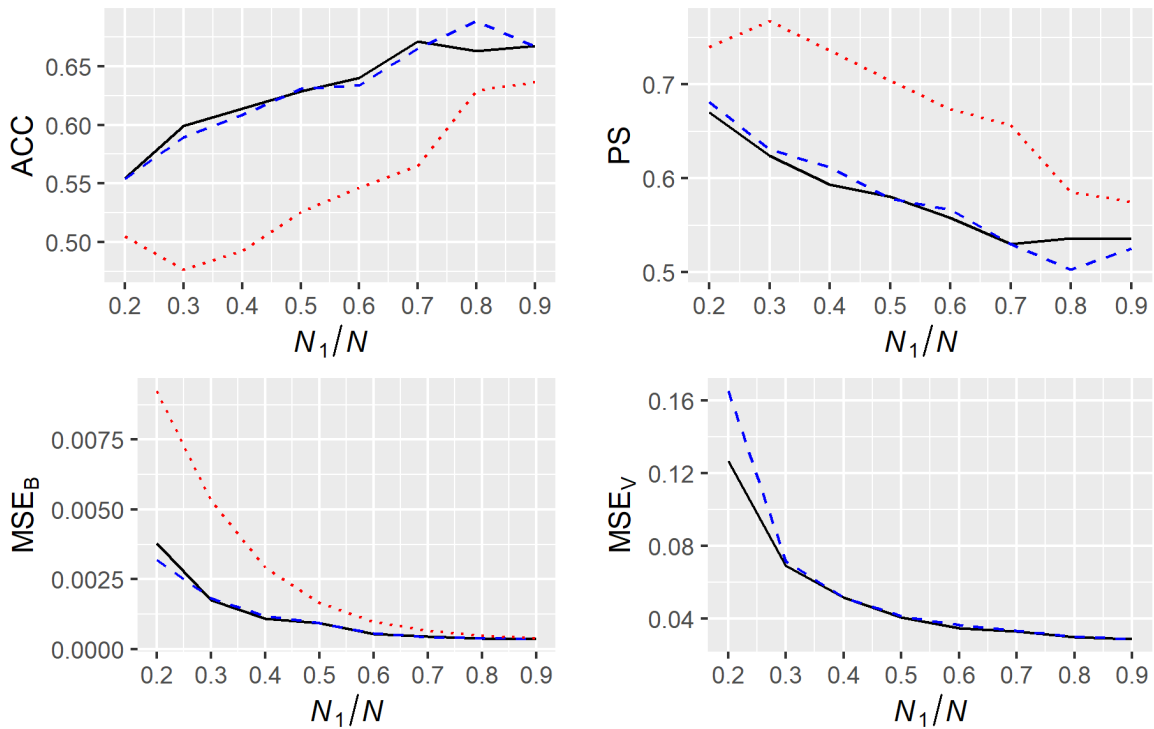

**Figure 2.** Similar to Figure 1 ($N = 100$), but with double noise levels: $\xi_1 = 40\%$ and $\xi_2 = 10\%$.

It could be anticipated that the performances of all methods generally get better with larger sample sizes. To test this hypothesis, we redo the experiment ($\xi_1 = 40\%$ and $\xi_2 = 10\%$) with a larger sample size of $N = 500$. Figure 3 shows the resulting performances of the tested methods and confirms this hypothesis. In fact, all methods achieve even better performances than in the lower-noise case in Figure 1, thanks to having larger sample sizes. Interestingly, the proposed method produces quite robust ACC and PS performances to the $N_1/N$ ratio.

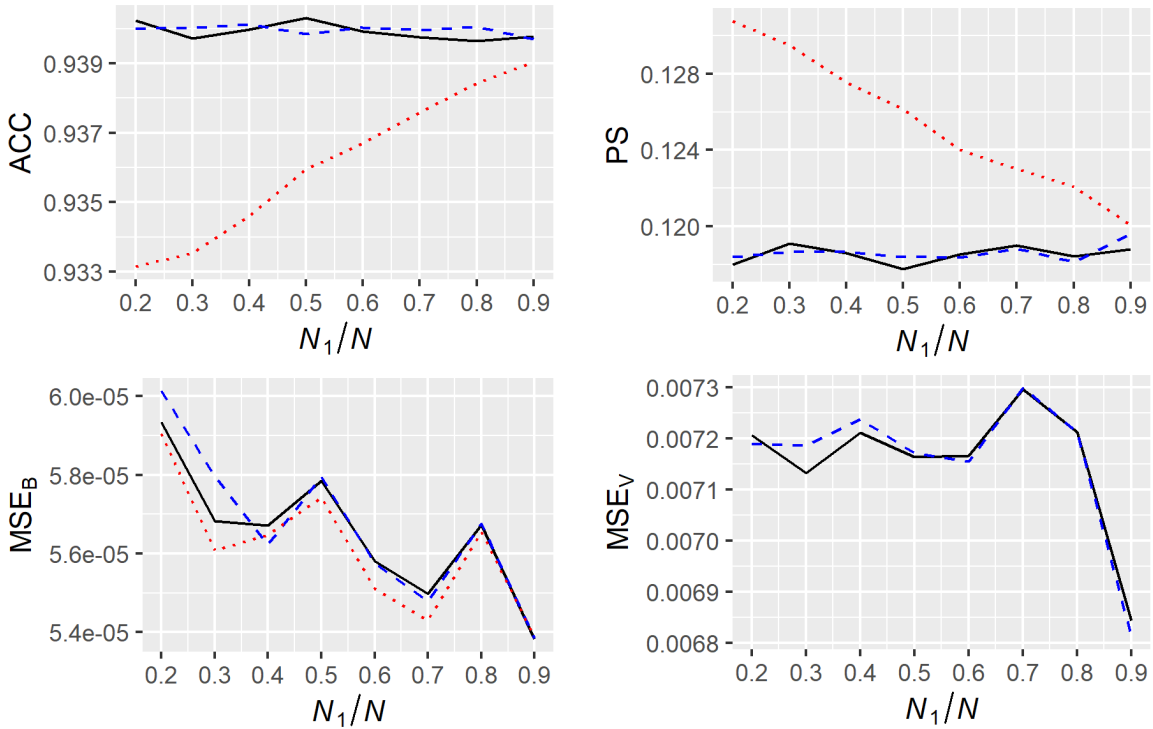

**Figure 3.** Similar to Figure 2 ($\xi_1 = 40\%$ and $\xi_2 = 10\%$), but with a larger sample size of $N = 500$.

To verify these observations with even larger noise levels, we now triple $\xi_1$ and $\xi_2$ to 60% and 15%, respectively. Figures 4 and 5 show the resulting performances for $N = 100$ and $N = 500$, respectively. We again see similar observations as above. First, the performances of all methods degrade as the noise levels increase, but this problem is alleviated by having larger sample sizes. Second, the proposed method is better than using only the complete data for conditional SMACOF (Bui 2024). It should be emphasized that this comparison does not account for the limitation of the latter approach, which cannot learn the coordinates of the objects with missing values in the reduced-dimension space. Third, the ACC and PS performances of the proposed method are robust to the $N_1/N$ ratio when $N = 500$. This suggests that users may not need all the known feature values to get the best dimension reduction result. We observe similar findings when using random weights for the seven car-brand features instead of the ones from Consumer Report (see Appendix F).

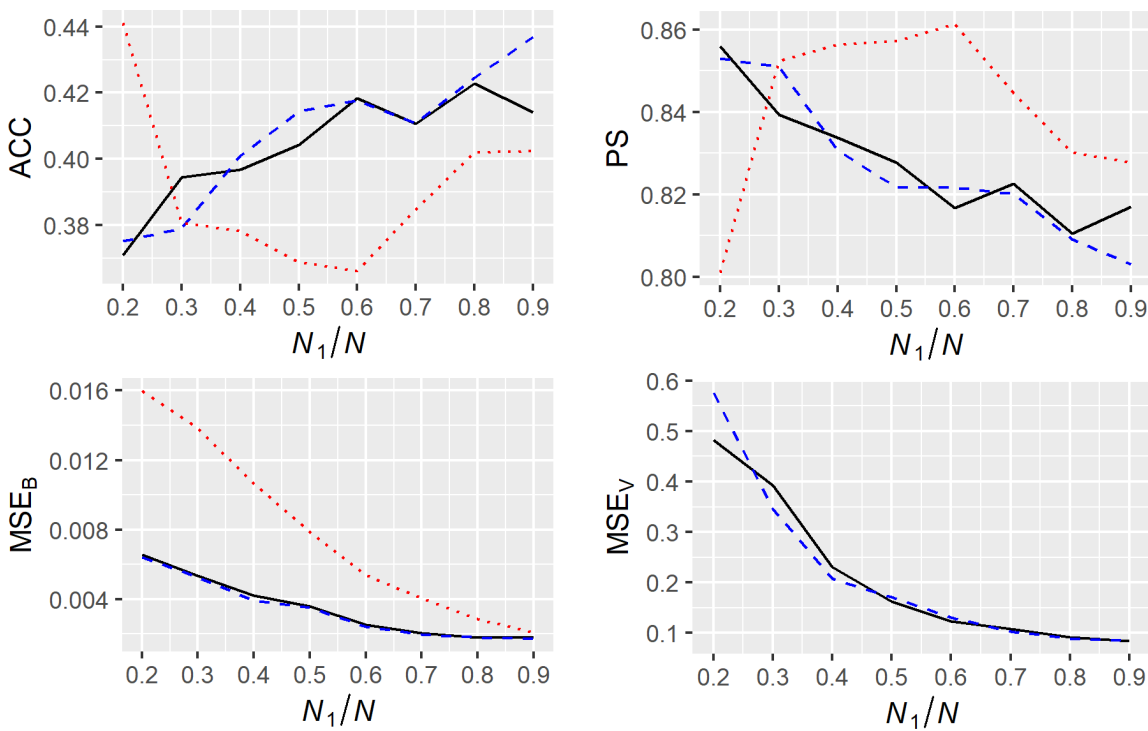

**Figure 4.** Similar to Figure 1 ($N = 100$), but with triple noise levels: $\xi_1 = 60\%$ and $\xi_2 = 15\%$.

### 4.2. Kinship Terms Example

This section tests the proposed method on the kinship terms dataset of the study of Rosenberg and Kim (1975). This dataset contains pairwise dissimilarities between 15 kinship terms, the name of which are shown in Table 2. These dissimilarities are the percentages of the times these 15 kinship terms were not grouped together by college students. The dataset also contains three variables: gender, generation, and kinship degree, which were concluded by Rosenberg and Kim to contribute mainly to the dissimilarities among the kinship terms. The values of these variables in Table 2 are provided in Mair et al. (2022).

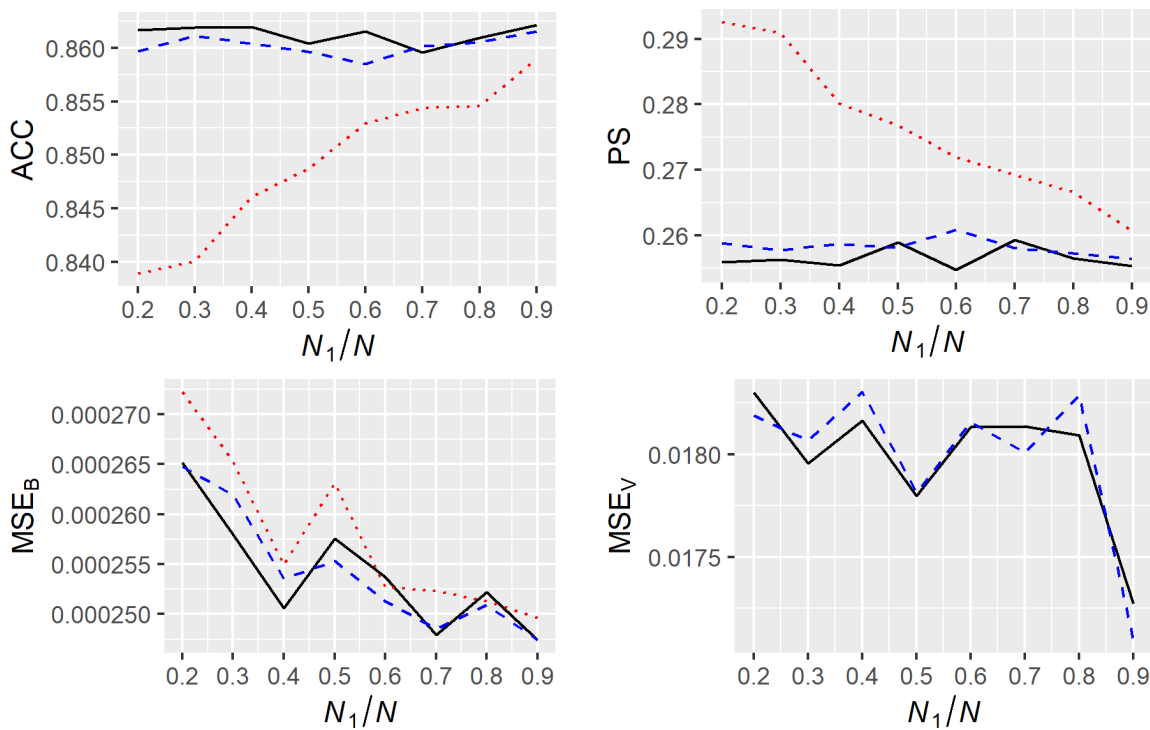

**Figure 5.** Similar to Figure 4 ($\xi_1 = 60\%$ and $\xi_2 = 15\%$), but with a larger sample size of $N = 500$.

**Table 2.** The variables that contribute the dissimilarities among the kinship terms found in the study of Rosenberg and Kim (1975).

|  | Grand-mother | Mother | Aunt | Sister | Daughter | Niece | Grand-daughter | Cousin | Grand-father | Father | Uncle | Brother | Son | Nephew | Grand-son |
|---|---|---|---|---|---|---|---|---|---|---|---|---|---|---|---|
| Gender | 2 | 2 | 2 | 2 | 2 | 2 | 2 | NA | 1 | 1 | 1 | 1 | 1 | 1 | 1 |
| Generation | -2 | -1 | -1 | 0 | 1 | 1 | 2 | 0 | -2 | -1 | -1 | 0 | 1 | 1 | 2 |
| Degree | 2 | 1 | 3 | 2 | 1 | 3 | 2 | 4 | 2 | 1 | 3 | 2 | 1 | 3 | 2 |

We apply the proposed method to the pairwise dissimilarities between the kinship terms, using the gender variable as the known feature (see Appendix G for additional experiments with kinship degree and generation gap as the known feature). Note that the cousin term has a missing gender value due to its gender ambiguity. We use the method in Bui (2022) to initialize the proposed algorithm, as discussed in Section 3. Figure 6 plots the learned two-dimensional coordinates of the kinship terms. From left to right of the plot (along Dimension 1), we can see an increasing level of the kinship degrees that match quite well with the values in Table 2 (which are better understood as ranks, not continuous values). Similarly, from top to bottom of the plot (along Dimension 2), we can see an increasing trend of the generation gap of the kinship terms (the absolute value of the generation variable in Table 2). These findings agree with Rosenberg and Kim (1975) that kinship degree and generation contribute to the dissimilarities between the kinship terms.

The imputed gender value for the cousin term by the proposed method is 1.493 (see Remark 2 for how the imputation can be done). This shows the balanced view of the college students in the study of Rosenberg and Kim (1975) for the gender of the cousin term, which is coded in their judgement of the dissimilarities between the 15 kinship terms. Thus, the ability to impute missing known feature values of the proposed method can also provide additional insights into the problem.

**Remark 5:** Figure 6 contains a somewhat counter-intuitive result: Aunt/Uncle is closer to Sister/Brother than Niece/Nephew is. A potential explanation for this issue is that the solution obtained in Figure 6 is not the best possible. Applying the proposed algorithm using the method in Bui (2024) for initialization with different seeds of the random number generator, we obtain the solution in Figure 7, which has the lowest normalized conditional stress value of 0.0260 < 0.0264 of the solution in Figure 6. Same with Figure 6, the kinship degree and generation gap features can also be seen in Figure 7. Furthermore, the imputed value for Gender of Cousin is 1.437 in this solution, which is close to that in the solution of Figure 6. However, the counter-intuitive result mentioned above does not occur in Figure 7. In general, a solution with a lower normalized conditional stress value is preferred because it fits the observed data better. Hence, users are recommended to trial different initializations to find the best possible solution.

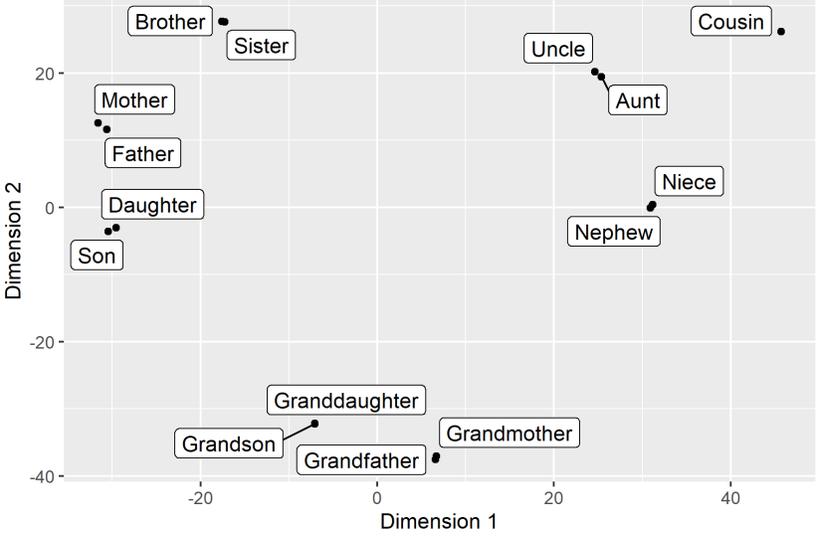

**Figure 6.** Plotting the two-dimensional coordinates of the 15 kinship terms learned by the proposed method, using Gender as the known feature. Along Dimension 1, there is a transition in the level of the kinship degrees. Along Dimension 2, there is a transition in the generation gap of the terms.

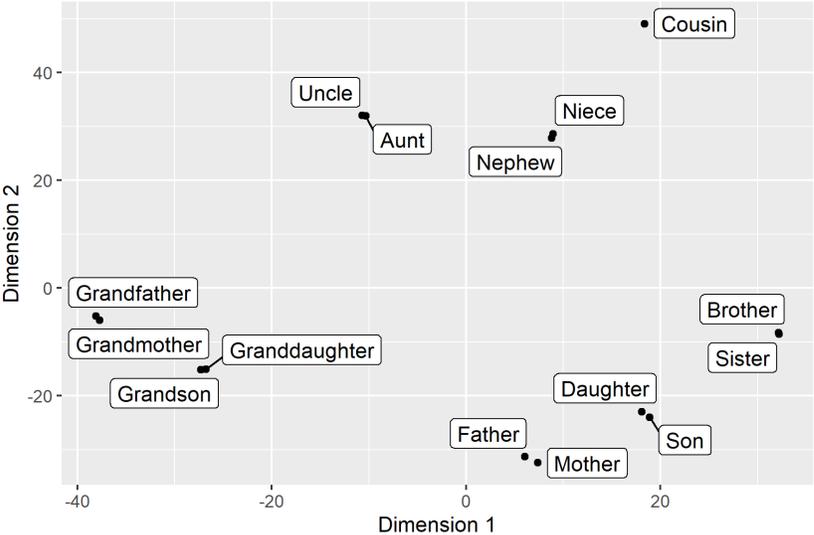

**Figure 7.** Plotting a similar result to Figure 6, but for a solution corresponding to a lower normalized conditional stress value (0.0260 < 0.0264 in Figure 6).

## 5. Conclusions

To be applicable, existing conditional multidimensional scaling approaches need to discard objects with missing data in the known features. This is undesirable when the coordinates of the discarded objects in the reduced-dimension space are of interest to practitioners. Furthermore, discarding data generally leads to poorer estimation of the low-dimensional configuration. To address this limitation, this paper presents a conditional multidimensional scaling method that can also learn coordinates of the objects with missing known feature values in the reduced-dimension space and impute the missing values.

The proposed method is tested on a simulated car-brand perception example and a real kinship term example. These examples demonstrate that the proposed method can improve the estimation quality of the low-dimensional configuration. The estimation quality remains high even when the ratio of missing values in the known features is large. This allows users to collect incomplete information of the known features intentionally to reduce time and cost. Moreover, the ability to impute missing known feature values can provide additional insights from the data.

Expectation-maximum algorithms are commonly used when there are missing data. It would be interesting to consider such an approach for the problem in this paper. However, a thorough study is warranted to develop such a method and compare it with the proposed method in this paper. Therefore, this research direction will be left for future studies.

**Appendix**

**A. A Closed-Form Solution for Conditional MDS**

The closed-form solution for conditional MDS by Bui (2022) is briefly reviewed here. This solution consists of two main steps. Step 1 estimates $\mathbf{B}$, and Step 2 estimates $\mathbf{U}$ based on the estimate of $\mathbf{B}$ obtained in Step 1.

First, denote $\mathbf{R}$ as the $q \times q$ PCA whitening matrix of $\mathbf{V}$ and $\mathbf{r}_k^T$ ($k = 1, \ldots, q$) as its rows. $\mathbf{R}$ is assumed to be full rank (i.e., $\mathbf{VB} = \mathbf{VRR}^{-1}\mathbf{B}$). This assumption can be easily satisfied by having only linearly independent columns in $\mathbf{V}$. Step 1 estimates $\mathbf{B}$ by $\mathbf{B} = \mathbf{R}\,\text{diag}(\beta_1^{1/2}, \ldots, \beta_q^{1/2})$, where the $\beta_k$'s are coefficients of the following linear regression model:

$$\delta_{ij}^2 = \mu + \sum_{k=1}^{q} \beta_k \left[\mathbf{r}_k^T(\mathbf{v}_i - \mathbf{v}_j)\right]^2 + \epsilon_{ij}, \quad 1 \leq i < j \leq N.$$

This estimate results in $\delta_{ij}^2 \approx d_{ij}^2(\mathbf{U}, \mathbf{B})$, $1 \leq i < j \leq N$ (see Bui (2022) for details). This approach assumes that, in expectation, the $\delta_{ij}^2$'s do not decrease as the $\left[\mathbf{r}_k^T(\mathbf{v}_i - \mathbf{v}_j)\right]^2$'s increase for $\forall k$. This assumption generally holds, unless the $\delta_{ij}^2$'s behave strangely. Nevertheless, some estimate of the $\beta_k$'s (e.g., from ordinary least squares) can be negative due to estimation uncertainty. A quick fix for this problem is setting negative coefficient estimates to 0. Methods for producing non-negative coefficient estimates are available, but they may not necessarily give closed-form solutions.

Second, denote $\mathbf{A} = [a_{ij}]_{i,j=1,\ldots,N}$, where $a_{ij} = -\frac{1}{2}\delta_{ij}^2$, $\mathbf{M} = (\mathbf{I}_N - N^{-1}\mathbf{1}_{N \times N})$, where $\mathbf{I}_N$ is an $N \times N$ identity matrix and $\mathbf{1}_{N \times N}$ is an $N \times N$ matrix with all elements equal to 1. Bui (2022) showed that if the given dissimilarities are Euclidean, then $\mathbf{MUU}^T\mathbf{M} = \mathbf{M}(\mathbf{A} - \mathbf{VBB}^T\mathbf{V}^T)\mathbf{M}$, which is positive semidefinite. Let $\mathbf{\Lambda}_p^{1/2}$ be the $p \times p$ diagonal matrix of the square roots of the $p$ largest eigenvalues of $\mathbf{M}(\mathbf{A} - \mathbf{VBB}^T\mathbf{V}^T)\mathbf{M}$. Additionally, let $\mathbf{E}_p$ be the $N \times p$ matrix that contains the $p$ eigenvectors corresponding to the $p$ largest eigenvalues. Step 2 of the solution by Bui (2022) estimates "centered" $\mathbf{U}$ via $\hat{\mathbf{U}} = \mathbf{E}_p \mathbf{\Lambda}_p^{1/2}$, which may yield a good approximation even when the given dissimilarities are non-Euclidean.

## B. Lemma 1

*Under Assumptions 1(a) and 1(b):*

a) $\mathbf{H}_{11}$ *and* $\mathbf{H}_{22}$ *are Stieltjes matrices (i.e., they are real symmetric positive definite matrices).*

b) $\mathbf{H}$ *is positive semi-definite.*

This lemma is needed for proving Theorem 1.

**Proof:**

a) We will first prove that $\mathbf{H}_{11}$ is positive definite. Consider an arbitrary vector $\mathbf{a}^T = [a_1, a_2, \ldots, a_{N_1}]^T \neq \mathbf{0}$, it can be verified that under Assumption 1(a):

$$\mathbf{a}^T \mathbf{H}_{11} \mathbf{a} = \sum_{j=1}^{N_1} \sum_{k=1}^{N_1} w_{jk} a_j (a_j - a_k) + \sum_{j=1}^{N_1} \sum_{k=N_1+1}^{N} w_{jk} a_j^2$$

$$= \sum_{1 \leq j < k \leq N_1} w_{jk} (a_j - a_k)^2 + \sum_{j=1}^{N_1} \sum_{k=N_1+1}^{N} w_{jk} a_j^2. \tag{11}$$

Under Assumption 1(b), there exists at least a positive $w_{jk}$ for each object $j = 1, 2, \ldots, N_1$ and an object $k > j$. As a result, the first summand in (11) is non-negative, and the second summand in (11) is positive. Therefore, $\mathbf{H}_{11}$ is positive definite, and so is $\mathbf{H}_{22}$ by permutation. An alternative proof can be obtained by noting that $\mathbf{H}_{11}$ and $\mathbf{H}_{22}$ are Z-matrices with positive row sums. This class of matrices has been shown to have positive determinant by Hermann Minkowski (Bermon and Plemmons 1994). By definition, real symmetric positive definite matrices $\mathbf{H}_{11}$ and $\mathbf{H}_{22}$ are Stieltjes matrices∎

b) When $N_1 = N$, it can be shown that (11) reduces to $\mathbf{a}^T \mathbf{H} \mathbf{a} = \sum_{1 \leq j < k \leq N} w_{jk} (a_j - a_k)^2$, which is non-negative for any arbitrary vector $\mathbf{a}^T = [a_1, a_2, \ldots, a_{N_1}]^T \neq \mathbf{0}$. Thus, $\mathbf{H}$ is positive semi-definite∎

## C. Lemma 2

*Under Assumption 1, $\mathbf{S}$, $\mathbf{K}_{\tilde{v}_2}$, and $\mathbf{K}_b$ are invertible.*

This lemma is needed for proving Theorem 1.

**Proof:**

- $\mathbf{S}$ is invertible:

We will prove that $\mathbf{S} = \mathbf{V}_1^T \mathbf{H}_{11} \mathbf{V}_1$ is invertible by contradiction. As noted in Bui (2024), $\sum_{1 \leq j < k \leq N} w_{jk}(v_{jt} - v_{kt})(v_{jm} - v_{km})$ is the value at the *t*th row and *m*th column of $\mathbf{V}^T \mathbf{H} \mathbf{V}$. Suppose to the contrary that $\mathbf{S}$ is not invertible, then there exists a vector $[a_1, a_2, \ldots, a_q]^T \neq \mathbf{0}$ such that

$\sum_{m=1}^{q} a_m \sum_{j<k} w_{jk}(v_{jt} - v_{kt})(v_{jm} - v_{km}) = 0$ for each $t = 1, 2, \ldots, N_1$

$\rightarrow \sum_{t=1}^{q} a_t \sum_{j<k} w_{jk}(v_{jt} - v_{kt}) \sum_{m=1}^{q} a_m(v_{jm} - v_{km}) = 0$

$\leftrightarrow \sum_{j<k} w_{jk} \sum_{t=1}^{q} a_t(v_{jt} - v_{kt}) \sum_{m=1}^{q} a_m(v_{jm} - v_{km}) = 0$

$\leftrightarrow \sum_{j<k} w_{jk} \left[ \sum_{m=1}^{q} a_m(v_{jm} - v_{km}) \right]^2 = 0.$

Under Assumption 1(b), there exists at least a positive $w_{jk}$ for each object $j = 1, 2, \ldots, N_1$ and an object $k > j$. As a result, $\sum_{m=1}^{q} a_m(v_{jm} - v_{km}) = 0$ for $1 \leq j < k \leq N_1$. However, Assumption 1(c) implies that there exist $q$ pairs of $(j, k)$'s such that $\sum_{m=1}^{q} a_m(v_{jm} - v_{km}) = 0$ if and only if $[a_1, a_2, \ldots, a_q]^T = \mathbf{0}$. Therefore, $\mathbf{S}$ is invertible by contradiction ∎

- $\mathbf{K}_{\tilde{v}_2}$ is invertible:

Note that $\det(-\mathbf{K}_{\tilde{v}_2}) = \det(-\mathbf{H}_{22} + \mathbf{H}_{21}\mathbf{V}_1 \mathbf{S}^{-1} \mathbf{V}_1^T \mathbf{H}_{12})$. Based on Lemmas 1(a) and 2, we have that $-\mathbf{H}_{22}$ and $\mathbf{V}_1^T \mathbf{H}_{11} \mathbf{V}_1$ are invertible under Assumption 1. Hence, it follows from the matrix determinant lemma (Harville 1997) that

$\det(-\mathbf{K}_{\tilde{v}_2}) = \det(-\mathbf{H}_{22}) \det(\mathbf{I}_{N_1} - \mathbf{S}^{-1/2} \mathbf{V}_1^T \mathbf{H}_{12} \mathbf{H}_{22}^{-1} \mathbf{H}_{21} \mathbf{V}_1 \mathbf{S}^{-1/2}),$

where $\mathbf{I}_{N_1}$ is an $N_1 \times N_1$ identity matrix. We will first show that the eigenvalues of $\mathbf{S}^{-1/2}\mathbf{V}_1^T\mathbf{H}_{12}\mathbf{H}_{22}^{-1}\mathbf{H}_{21}\mathbf{V}_1\mathbf{S}^{-1/2}$ are less than 1 via contradiction. Specifically, suppose to the contrary that these eigenvalues are not less than 1, then

$$\mathbf{S}^{-1/2}\mathbf{V}_1^T\mathbf{H}_{12}\mathbf{H}_{22}^{-1}\mathbf{H}_{21}\mathbf{V}_1\mathbf{S}^{-1/2} \succcurlyeq \mathbf{I}_{N_1}$$

$$\rightarrow \mathbf{V}_1^T\mathbf{H}_{12}\mathbf{H}_{22}^{-1}\mathbf{H}_{21}\mathbf{V}_1 \succcurlyeq \mathbf{S}$$

$$\rightarrow \mathbf{V}_1^T(\mathbf{H}_{12}\mathbf{H}_{22}^{-1}\mathbf{H}_{21} - \mathbf{H}_{11})\mathbf{V}_1 \succcurlyeq \mathbf{0}$$

$$\rightarrow \mathbf{H}_{12}\mathbf{H}_{22}^{-1}\mathbf{H}_{21} - \mathbf{H}_{11} \succcurlyeq \mathbf{0}$$

However, $\mathbf{H}_{11} - \mathbf{H}_{12}\mathbf{H}_{22}^{-1}\mathbf{H}_{21} \succcurlyeq \mathbf{0}$ because it is the Schur complement of the positive definite block $\mathbf{H}_{22}$ of the positive semi-definite matrix $\mathbf{H}$ (Lemma 1). As a result, $\mathbf{H}_{11} = \mathbf{H}_{12}\mathbf{H}_{22}^{-1}\mathbf{H}_{21}$. This is not true as at least some off-diagonal elements of $\mathbf{H}_{11}$ is negative, whereas $\mathbf{H}_{12}\mathbf{H}_{22}^{-1}\mathbf{H}_{21}$ can be shown to be a non-negative matrix (proof sketch: $\mathbf{H}_{21}$ is a non-positive matrix; $\mathbf{H}_{22}^{-1}$ is a non-negative matrix because $\mathbf{H}_{22}$ is a Stieltjes matrix as shown in Lemma 1). Therefore, the eigenvalues of $\mathbf{S}^{-1/2}\mathbf{V}_1^T\mathbf{H}_{12}\mathbf{H}_{22}^{-1}\mathbf{H}_{21}\mathbf{V}_1\mathbf{S}^{-1/2}$ are less than 1 by contradition. This means that $\det(\mathbf{I}_{N_1} - \mathbf{S}^{-1/2}\mathbf{V}_1^T\mathbf{H}_{12}\mathbf{H}_{22}^{-1}\mathbf{H}_{21}\mathbf{V}_1\mathbf{S}^{-1/2}) \neq 0$, which also means that $\det(\mathbf{K}_{\tilde{v}_2}) \neq 0$. Therefore, $\mathbf{K}_{\tilde{v}_2}$ is invertible∎

- $\mathbf{K}_b$ is invertible:

Again, from the generalized matrix determinant lemma:

$$\det(-\mathbf{K}_{\tilde{v}_2}) = \det(-\mathbf{H}_{22})\det(\mathbf{S}^{-1})\det(\mathbf{S} - \mathbf{V}_1^T\mathbf{H}_{12}\mathbf{H}_{22}^{-1}\mathbf{H}_{21}\mathbf{V}_1)$$

$$= \det(-\mathbf{H}_{22})\det(\mathbf{S}^{-1})\det(\mathbf{K}_b).$$

From Part (a), $\det(\mathbf{K}_{\tilde{v}_2}) \neq 0$, and therefore, $\det(\mathbf{K}_b) \neq 0$. As such, $\mathbf{K}_b$ is invertible∎

## D. Proof of Theorem 1

Assume that Assumption 1 holds and expand (1) to:

$$\sigma(\mathbf{U}, \widetilde{\mathbf{V}}) = \eta_{\tilde{\delta}}^2 + \eta^2(\mathbf{U}, \widetilde{\mathbf{V}}) - 2\rho(\mathbf{U}, \widetilde{\mathbf{V}}), \tag{12}$$

where $\eta_{\tilde{\delta}}^2 = \sum_{i<j} w_{ij} \tilde{\delta}_{ij}^2$ is a constant, $\eta^2(\mathbf{U}, \widetilde{\mathbf{V}}) = \sum_{i<j} w_{ij} d_{ij}^2(\mathbf{U}, \widetilde{\mathbf{V}})$, and $\rho(\mathbf{U}, \widetilde{\mathbf{V}}) = \sum_{i<j} w_{ij} \tilde{\delta}_{ij} d_{ij}(\mathbf{U}, \widetilde{\mathbf{V}})$. It can be shown that $\eta^2(\mathbf{U}, \widetilde{\mathbf{V}}) = \text{tr}(\mathbf{U}^T \mathbf{H} \mathbf{U}) + \text{tr}(\mathbf{B}^T \mathbf{V}^T \mathbf{H} \mathbf{V} \mathbf{B})$. Hence,

$$\eta^2(\mathbf{U}, \widetilde{\mathbf{V}}) = \text{tr}\mathbf{U}^T \mathbf{H} \mathbf{U} + \text{tr}[\mathbf{B}^T \mathbf{V}_1^T, \widetilde{\mathbf{V}}_2^T] \begin{bmatrix} \mathbf{H}_{11} & \mathbf{H}_{12} \\ \mathbf{H}_{21} & \mathbf{H}_{22} \end{bmatrix} \begin{bmatrix} \mathbf{V}_1 \mathbf{B} \\ \widetilde{\mathbf{V}}_2 \end{bmatrix}$$

$$= \mathbf{U}^T \mathbf{H} \mathbf{U} + \text{tr}\mathbf{B}^T \mathbf{S} \mathbf{B} + 2\text{tr}\mathbf{B}^T \mathbf{V}_1^T \mathbf{H}_{12} \widetilde{\mathbf{V}}_2 + \text{tr}\widetilde{\mathbf{V}}_2^T \mathbf{H}_{22} \widetilde{\mathbf{V}}_2. \tag{13}$$

Let $\mathbf{Z}_u \in \mathbb{R}^{N \times p}$, $\mathbf{Z}_b \in \mathbb{R}^{q \times q}$, and $\mathbf{Z}_{\tilde{v}_2} \in \mathbb{R}^{(N-N_1) \times q}$ be matrices with the same shape as that of $\mathbf{U}$, $\mathbf{B}$, and $\widetilde{\mathbf{V}}_2$, respectively. As in Bui (2024), we have that

$$-\rho(\mathbf{U}, \widetilde{\mathbf{V}}) \leq -\text{tr}\left(\begin{bmatrix} \mathbf{U} & \mathbf{V}_1 \mathbf{B} \\ & \widetilde{\mathbf{V}}_2 \end{bmatrix}^T \mathbf{C} \begin{bmatrix} \mathbf{Z}_u & \mathbf{V}_1 \mathbf{Z}_b \\ & \mathbf{Z}_{\tilde{v}_2} \end{bmatrix}\right)$$

$$= -\text{tr}\mathbf{U}^T \mathbf{C} \mathbf{Z}_u - \text{tr}\mathbf{B}^T \mathbf{V}_1^T \mathbf{C}_{11} \mathbf{V}_1 \mathbf{Z}_b - \text{tr}\mathbf{B}^T \mathbf{V}_1^T \mathbf{C}_{12} \mathbf{Z}_{\tilde{v}_2} - \text{tr}\widetilde{\mathbf{V}}_2^T \mathbf{C}_{21} \mathbf{V}_1 \mathbf{Z}_b - \text{tr}\widetilde{\mathbf{V}}_2^T \mathbf{C}_{22} \mathbf{Z}_{\tilde{v}_2}, \tag{14}$$

Plugging (13) and (14) into (12), we have

$$\sigma(\mathbf{U}, \widetilde{\mathbf{V}}) \leq \eta_{\tilde{\delta}}^2 + \text{tr}(\mathbf{U}^T \mathbf{H} \mathbf{U}) + \text{tr}\mathbf{B}^T \mathbf{S} \mathbf{B} + 2\text{tr}\mathbf{B}^T \mathbf{V}_1^T \mathbf{H}_{12} \widetilde{\mathbf{V}}_2 + \text{tr}\widetilde{\mathbf{V}}_2^T \mathbf{H}_{22} \widetilde{\mathbf{V}}_2 - 2\text{tr}\mathbf{U}^T \mathbf{C} \mathbf{Z}_u -$$

$$2\text{tr}\mathbf{B}^T \mathbf{V}_1^T \mathbf{C}_{11} \mathbf{V}_1 \mathbf{Z}_b - 2\text{tr}\mathbf{B}^T \mathbf{V}_1^T \mathbf{C}_{12} \mathbf{Z}_{\tilde{v}_2} - 2\text{tr}\widetilde{\mathbf{V}}_2^T \mathbf{C}_{21} \mathbf{V}_1 \mathbf{Z}_b - 2\text{tr}\widetilde{\mathbf{V}}_2^T \mathbf{C}_{22} \mathbf{Z}_{\tilde{v}_2}$$

$$= \tau(\mathbf{U}, \mathbf{B}, \widetilde{\mathbf{V}}_2, \mathbf{Z}_u, \mathbf{Z}_b, \mathbf{Z}_{\tilde{v}_2}),$$

which is a majorizing function of $\sigma(\cdot)$. This function has a quadratic form in $\mathbf{U}$, $\mathbf{B}$, and $\widetilde{\mathbf{V}}_2$; therefore, its global minimizer over $\mathbf{U}$, $\mathbf{B}$, and $\widetilde{\mathbf{V}}_2$ satisfies:

$$\begin{cases} \mathbf{0} = \frac{\partial \tau(\mathbf{U}, \mathbf{B}, \widetilde{\mathbf{V}}_2, \mathbf{Z}_u, \mathbf{Z}_b, \mathbf{Z}_{\tilde{v}_2})}{\partial \mathbf{U}} = 2\mathbf{H}\mathbf{U} - 2\mathbf{C}\mathbf{Z}_u \\ \mathbf{0} = \frac{\partial \tau(\mathbf{U}, \mathbf{B}, \widetilde{\mathbf{V}}_2, \mathbf{Z}_u, \mathbf{Z}_b, \mathbf{Z}_{\tilde{v}_2})}{\partial \mathbf{B}} = 2\mathbf{S}\mathbf{B} + 2\mathbf{V}_1^T \mathbf{H}_{12} \widetilde{\mathbf{V}}_2 - 2\mathbf{V}_1^T \mathbf{C}_{11} \mathbf{V}_1 \mathbf{Z}_b - 2\mathbf{V}_1^T \mathbf{C}_{12} \mathbf{Z}_{\tilde{v}_2} \\ \mathbf{0} = \frac{\partial \tau(\mathbf{U}, \mathbf{B}, \widetilde{\mathbf{V}}_2, \mathbf{Z}_u, \mathbf{Z}_b, \mathbf{Z}_{\tilde{v}_2})}{\partial \widetilde{\mathbf{V}}_2} = 2\mathbf{H}_{21} \mathbf{V}_1 \mathbf{B} + 2\mathbf{H}_{22} \widetilde{\mathbf{V}}_2 - 2\mathbf{C}_{21} \mathbf{V}_1 \mathbf{Z}_b - 2\mathbf{C}_{22} \mathbf{Z}_{\tilde{v}_2} \end{cases}$$

$$\leftrightarrow \begin{cases} \mathbf{HU} = \mathbf{CZ}_u \\ \mathbf{SB} = \mathbf{V}_1^T \mathbf{C}_{11} \mathbf{V}_1 \mathbf{Z}_b + \mathbf{V}_1^T \mathbf{C}_{12} \mathbf{Z}_{\tilde{v}_2} - \mathbf{V}_1^T \mathbf{H}_{12} \widetilde{\mathbf{V}}_2 \\ \mathbf{H}_{22} \widetilde{\mathbf{V}}_2 = \mathbf{C}_{21} \mathbf{V}_1 \mathbf{Z}_b + \mathbf{C}_{22} \mathbf{Z}_{\tilde{v}_2} - \mathbf{H}_{21} \mathbf{V}_1 \mathbf{B} \end{cases} \quad (15)$$

Thus, a minimizer of $\tau(\cdot)$ over $\mathbf{U}$ can be obtained via the Guttman transform (Guttman 1968):

$$\mathbf{U} = \mathbf{H}^+ \mathbf{CZ}_u. \quad (16)$$

And the minimizers of $\tau(\cdot)$ over $\mathbf{B}$ and $\widetilde{\mathbf{V}}_2$ satisfy:

$$\begin{cases} \mathbf{B} = \mathbf{S}^{-1} \mathbf{V}_1^T \big( \mathbf{C}_{11} \mathbf{V}_1 \mathbf{Z}_b + \mathbf{C}_{12} \mathbf{Z}_{\tilde{v}_2} - \mathbf{H}_{12} \widetilde{\mathbf{V}}_2 \big) \\ \widetilde{\mathbf{V}}_2 = \mathbf{H}_{22}^{-1} \big( \mathbf{C}_{21} \mathbf{V}_1 \mathbf{Z}_b + \mathbf{C}_{22} \mathbf{Z}_{\tilde{v}_2} - \mathbf{H}_{21} \mathbf{V}_1 \mathbf{B} \big) \end{cases}, \quad (17)$$

as $\mathbf{H}_{22}$ and $\mathbf{S}$ are invertible by Lemmas 1 and 2, respectively. Plugging (17) into (15), we have

$$\begin{cases} \mathbf{SB} = \mathbf{V}_1^T \mathbf{C}_{11} \mathbf{V}_1 \mathbf{Z}_b + \mathbf{V}_1^T \mathbf{C}_{21}^T \mathbf{Z}_{\tilde{v}_2} - \mathbf{V}_1^T \mathbf{H}_{12} \mathbf{H}_{22}^{-1} \big( \mathbf{C}_{21} \mathbf{V}_1 \mathbf{Z}_b + \mathbf{C}_{22} \mathbf{Z}_{\tilde{v}_2} - \mathbf{H}_{21} \mathbf{V}_1 \mathbf{B} \big) \\ \mathbf{H}_{22} \widetilde{\mathbf{V}}_2 = \mathbf{C}_{21} \mathbf{V}_1 \mathbf{Z}_b + \mathbf{C}_{22} \mathbf{Z}_{\tilde{v}_2} - \mathbf{H}_{21} \mathbf{V}_1 \mathbf{S}^{-1} \big( \mathbf{V}_1^T \mathbf{C}_{11} \mathbf{V}_1 \mathbf{Z}_b + \mathbf{V}_1^T \mathbf{C}_{12} \mathbf{Z}_{\tilde{v}_2} - \mathbf{V}_1^T \mathbf{H}_{12} \widetilde{\mathbf{V}}_2 \big) \end{cases}$$

$$\rightarrow \begin{cases} \mathbf{K}_b \mathbf{B} = \big( \mathbf{V}_1^T \mathbf{C}_{11} \mathbf{V}_1 - \mathbf{V}_1^T \mathbf{H}_{12} \mathbf{H}_{22}^{-1} \mathbf{C}_{21} \mathbf{V}_1 \big) \mathbf{Z}_b + \big( \mathbf{V}_1^T \mathbf{C}_{12} - \mathbf{V}_1^T \mathbf{H}_{12} \mathbf{H}_{22}^{-1} \mathbf{C}_{22} \big) \mathbf{Z}_{\tilde{v}_2} \\ \mathbf{K}_{\tilde{v}_2} \widetilde{\mathbf{V}}_2 = \big( \mathbf{C}_{21} \mathbf{V}_1 - \mathbf{G} \mathbf{V}_1^T \mathbf{C}_{11} \mathbf{V}_1 \big) \mathbf{Z}_b + \big( \mathbf{C}_{22} - \mathbf{G} \mathbf{V}_1^T \mathbf{C}_{12} \big) \mathbf{Z}_{\tilde{v}_2} \end{cases}$$

$$\rightarrow \begin{cases} \mathbf{B} = \mathbf{K}_b^{-1} \mathbf{V}_1^T \big[ \big( \mathbf{C}_{11} - \mathbf{H}_{12} \mathbf{H}_{22}^{-1} \mathbf{C}_{21} \big) \mathbf{V}_1 \mathbf{Z}_b + \big( \mathbf{C}_{12} - \mathbf{H}_{12} \mathbf{H}_{22}^{-1} \mathbf{C}_{22} \big) \mathbf{Z}_{\tilde{v}_2} \big] \\ \widetilde{\mathbf{V}}_2 = \mathbf{K}_{\tilde{v}_2}^{-1} \big[ \big( \mathbf{C}_{21} - \mathbf{G} \mathbf{V}_1^T \mathbf{C}_{11} \big) \mathbf{V}_1 \mathbf{Z}_b + \big( \mathbf{C}_{22} - \mathbf{G} \mathbf{V}_1^T \mathbf{C}_{12} \big) \mathbf{Z}_{\tilde{v}_2} \big] \end{cases} \quad (18)$$

because $\mathbf{K}_b$ and $\mathbf{K}_{\tilde{v}_2}$ are invertible by Lemma 2.

Then, the formulas (4—6) in Theorem 1 can be derived by setting $\mathbf{U} = \mathbf{U}^{[l]}$, $\mathbf{Z}_u = \mathbf{U}^{[l-1]}$, $\mathbf{B} = \mathbf{B}^{[l-1]}$, $\mathbf{Z}_b = \mathbf{B}^{[l-1]}$, $\widetilde{\mathbf{V}}_2 = \widetilde{\mathbf{V}}_2^{[l]}$, and $\mathbf{Z}_{\tilde{v}_2} = \widetilde{\mathbf{V}}_2^{[l-1]}$ in (16) and (18). Based on the properties of majorization and minimization, we have the following inequality chain: $\sigma\big(\mathbf{U}^{[l]}, \widetilde{\mathbf{V}}^{[l]}\big) \leq \tau\big(\mathbf{U}^{[l]}, \mathbf{B}^{[l]}, \widetilde{\mathbf{V}}_2^{[l]}, \mathbf{U}^{[l-1]}, \mathbf{B}^{[l-1]}, \widetilde{\mathbf{V}}_2^{[l-1]}\big) \leq \tau\big(\mathbf{U}^{[l-1]}, \mathbf{B}^{[l-1]}, \widetilde{\mathbf{V}}_2^{[l-1]}, \mathbf{U}^{[l-1]}, \mathbf{B}^{[l-1]}, \widetilde{\mathbf{V}}_2^{[l-1]}\big) = \sigma\big(\mathbf{U}^{[l-1]}, \widetilde{\mathbf{V}}^{[l-1]}\big)$, and the equality occurs if $\mathbf{U}^{[l]} = \mathbf{U}^{[l-1]}$, $\mathbf{B}^{[l]} = \mathbf{B}^{[l-1]}$, and $\widetilde{\mathbf{V}}_2^{[l]} = \widetilde{\mathbf{V}}_2^{[l-1]}$ ∎

### E. Theorem 2

*Let $\mathbf{I}_\bullet$ and $\mathbf{1}_{\bullet \times \bullet}$ be an identity matrix and a matrix of 1's, respectively, where the subscripts denote the sizes of these matrices. If all the weights $w_{ij}$'s are 1, we have the following identities:*

a) $\mathbf{H}^+ = \frac{1}{N}\left[\mathbf{I}_N - \frac{1}{N}\mathbf{1}_{N \times N}\right]$

b) $\mathbf{H}_{12}\mathbf{H}_{22}^{-1} = -\frac{1}{N_1}\mathbf{1}_{N_1 \times N_2}$

c) $\mathbf{S} = N\mathbf{V}_1^T\mathbf{V}_1 - \mathbf{v}_1^s[\mathbf{v}_1^s]^T$, where $\mathbf{v}_1^s$ is the vector of the column sums of $\mathbf{V}_1$

d) $\mathbf{S}^{-1} = \frac{1}{N}\left[\mathbf{I}_q + \frac{1}{N - \mathrm{tr}\left((\mathbf{V}_1^T\mathbf{V}_1)^{-1}\mathbf{v}_1^s[\mathbf{v}_1^s]^T\right)}(\mathbf{V}_1^T\mathbf{V}_1)^{-1}\mathbf{v}_1^s[\mathbf{v}_1^s]^T\right](\mathbf{V}_1^T\mathbf{V}_1)^{-1}$

e) $\mathbf{K}_b = N\mathbf{V}_1^T\mathbf{V}_1 - \frac{N}{N_1}\mathbf{v}_1^s[\mathbf{v}_1^s]^T$

f) $\mathbf{K}_b^{-1} = \frac{1}{N}\left[\mathbf{I}_q + \frac{1}{N_1 - \mathrm{tr}\left((\mathbf{V}_1^T\mathbf{V}_1)^{-1}\mathbf{v}_1^s[\mathbf{v}_1^s]^T\right)}(\mathbf{V}_1^T\mathbf{V}_1)^{-1}\mathbf{v}_1^s[\mathbf{v}_1^s]^T\right](\mathbf{V}_1^T\mathbf{V}_1)^{-1}$

g) $\mathbf{G}\mathbf{V}_1^T = -\mathbf{1}_{N_2 \times 1}([\mathbf{v}_1^s]^T\mathbf{S}^{-1}\mathbf{V}_1^T)$, i.e., $\mathbf{G}\mathbf{V}_1^T$ is a matrix with all rows equal to $[\mathbf{v}_1^s]^T\mathbf{S}^{-1}\mathbf{V}_1^T$

h) $\mathbf{K}_{\tilde{v}_2} = N\mathbf{I}_{N_2} - g\mathbf{1}_{N_2 \times N_2}$, where $g = 1 + [\mathbf{v}_1^s]^T\mathbf{S}^{-1}\mathbf{v}_1^s$

i) $\mathbf{K}_{\tilde{v}_2}^{-1} = \frac{1}{N}\left[\mathbf{I}_{N_2} + \frac{g}{N - gN_2}\mathbf{1}_{N_2 \times N_2}\right]$

This theorem shows how many formulas in the arbitrary weight cases can be simplified when the weights are equal. This theorem is the basis for Corollary 1.

**Proof:** When all the weights $w_{ij}$'s are 1, we have that $\mathbf{H} = N\mathbf{I}_N - \mathbf{1}_{N \times N}$, $\mathbf{H}_{11} = N\mathbf{I}_{N_1} - \mathbf{1}_{N_1 \times N_1}$, $\mathbf{H}_{22} = N\mathbf{I}_{N_2} - \mathbf{1}_{N_2 \times N_2}$, $\mathbf{H}_{21} = -\mathbf{1}_{N_2 \times N_1}$, and $\mathbf{H}_{12} = -\mathbf{1}_{N_1 \times N_2}$.

a) Using $\mathbf{H} = N\mathbf{I}_N - \mathbf{1}_{N \times N}$, we have that $(\mathbf{H} + \mathbf{1}_{N \times N})^{-1} = \frac{1}{N}\mathbf{I}_N$. Thus, the result follows from the fact that $\mathbf{H}^+ = (\mathbf{H} + \mathbf{1}_{N \times N})^{-1} - N^{-2}\mathbf{1}_{N \times N}$ (Borg and Groenen 2005, page 191) ∎

b) First, verify that $\mathbf{H}_{22}^{-1} = \frac{1}{N}\mathbf{I}_{N_2} + \frac{1}{NN_1}\mathbf{1}_{N_2 \times N_2}$ via the condition $\mathbf{H}_{22}^{-1}\mathbf{H}_{22} = \mathbf{I}_{N_2}$. It follows that

$\mathbf{H}_{12}\mathbf{H}_{22}^{-1} = -\mathbf{1}_{N_1 \times N_2}/N_1$ ∎

c) It can be verified that $\mathbf{V}_1^T \mathbf{1}_{N_1 \times N_1} \mathbf{V}_1 = \mathbf{v}_1^S [\mathbf{v}_1^S]^T$. As a result, $\mathbf{S} = \mathbf{V}_1^T \mathbf{H}_{11} \mathbf{V}_1 = \mathbf{V}_1^T (N \mathbf{I}_{N_1} - \mathbf{1}_{N_1 \times N_1}) \mathbf{V}_1 = N \mathbf{V}_1^T \mathbf{V}_1 - \mathbf{v}_1^S [\mathbf{v}_1^S]^T$ ∎

d) Because $\mathbf{V}_1^T \mathbf{V}_1$ is full-rank (which is implied by Assumption 1(c)) and $\mathbf{v}_1^S [\mathbf{v}_1^S]^T$ is a rank-1 matrix, the result follows from Equation 1 in Miller (1981) ∎

e) First, show that $\mathbf{H}_{12} \mathbf{H}_{22}^{-1} \mathbf{H}_{21} = \frac{N_2}{N_1} \mathbf{1}_{N_1 \times N_1}$, and then $\mathbf{V}_1^T \mathbf{H}_{12} \mathbf{H}_{22}^{-1} \mathbf{H}_{21} \mathbf{V}_1 = \mathbf{V}_1^T \frac{N_2}{N_1} \mathbf{1}_{N_1 \times N_1} \mathbf{V}_1 = \frac{N_2}{N_1} \mathbf{v}_1^S [\mathbf{v}_1^S]^T$. It follows that $\mathbf{K}_b = \mathbf{S} - \mathbf{V}_1^T \mathbf{H}_{12} \mathbf{H}_{22}^{-1} \mathbf{H}_{21} \mathbf{V}_1 = N \mathbf{V}_1^T \mathbf{V}_1 - \frac{N}{N_1} \mathbf{v}_1^S [\mathbf{v}_1^S]^T$ ∎

f) From Part (e) and similar to Part (d), the result follows from Equation 1 in Miller (1981) ∎

g) $\mathbf{H}_{21} \mathbf{V}_1$ is a matrix with all rows equal to $-[\mathbf{v}_1^S]^T$, i.e., $\mathbf{H}_{21} \mathbf{V}_1 = -\mathbf{1}_{N_2 \times 1} [\mathbf{v}_1^S]^T$. As a result, $\mathbf{G} = \mathbf{H}_{21} \mathbf{V}_1 \mathbf{S}^{-1} = -\mathbf{1}_{N_2 \times 1} [\mathbf{v}_1^S]^T \mathbf{S}^{-1}$, a matrix with all rows equal to $-[\mathbf{v}_1^S]^T \mathbf{S}^{-1}$. It follows that $\mathbf{G} \mathbf{V}_1^T = -\mathbf{1}_{N_2 \times 1} \left[ [\mathbf{v}_1^S]^T \mathbf{S}^{-1} \mathbf{v}_{1,1}, \dots, [\mathbf{v}_1^S]^T \mathbf{S}^{-1} \mathbf{v}_{1,N_1} \right] = -\mathbf{1}_{N_2 \times 1} \left[ \mathbf{v}_{1,1}^T \mathbf{S}^{-1} \mathbf{v}_1^S, \dots, \mathbf{v}_{1,N_1}^T \mathbf{S}^{-1} \mathbf{v}_1^S \right]$, a matrix with all rows equal to $-\left[ \mathbf{v}_{1,1}^T \mathbf{S}^{-1} \mathbf{v}_1^S, \dots, \mathbf{v}_{1,N_1}^T \mathbf{S}^{-1} \mathbf{v}_1^S \right]$ ∎

h) We have from the proof of Part (g) that $\mathbf{H}_{21} \mathbf{V}_1 = \mathbf{1}_{N_2 \times 1} [\mathbf{v}_1^S]^T$ and $\mathbf{G} = \mathbf{H}_{21} \mathbf{V}_1 \mathbf{S}^{-1} = \mathbf{1}_{N_2 \times 1} [\mathbf{v}_1^S]^T \mathbf{S}^{-1}$. Thus, $\mathbf{G} \mathbf{V}_1^T \mathbf{H}_{12} = \mathbf{1}_{N_2 \times 1} [\mathbf{v}_1^S]^T \mathbf{S}^{-1} \mathbf{v}_1^S \mathbf{1}_{1 \times N_2} = [\mathbf{v}_1^S]^T \mathbf{S}^{-1} \mathbf{v}_1^S \mathbf{1}_{N_2 \times N_2}$. Thus, $\mathbf{K}_{\tilde{v}_2} = \mathbf{H}_{22} - \mathbf{G} \mathbf{V}_1^T \mathbf{H}_{12} = N \mathbf{I}_{N_2} - \mathbf{1}_{N_2 \times N_2} - [\mathbf{v}_1^S]^T \mathbf{S}^{-1} \mathbf{v}_1^S \mathbf{1}_{N_2 \times N_2} = N \mathbf{I}_{N_2} - g \mathbf{1}_{N_2 \times N_2}$.

i) We can then verify the result via Part (h) and the condition $K_{\tilde{v}_2}^{-1} K_{\tilde{v}_2} = \mathbf{I}_{N_2}$ ∎

**F. Additional Experiments for the Car-Brand Perception Simulation Example**

Instead of fixing the weights of the seven car-brand features to those in the 2014 Car-Brand Perception Survey of Consumer Reports, here we randomly generate the weights in each Monte Carlo replicate, as follows. We first randomly sample seven Uniform(3, 7) values. Then, we divide them by their sum so that the resulting weights (ranging from 1/15 to 7/25) sum up to 1. Figures 8—12 show the performances of the tested methods using the same settings as in Section 4.1.

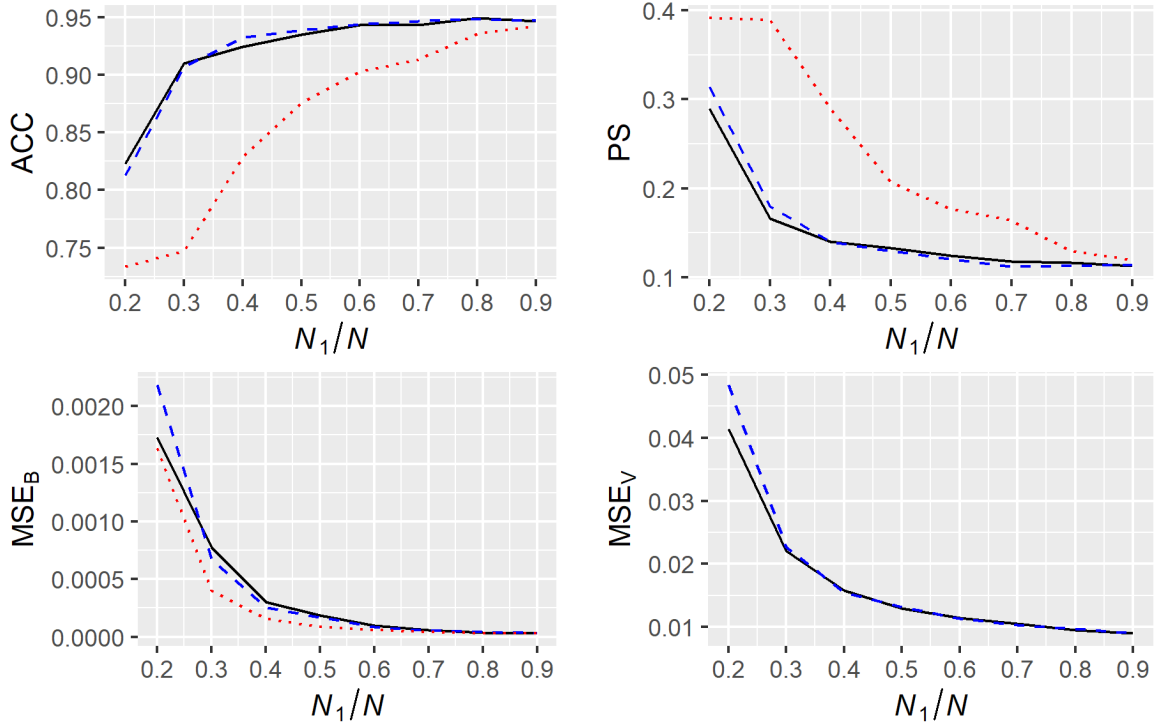

**Figure 8.** Similar to Figure 1 ($N = 100$, $\xi_1 = 20\%$, and $\xi_2 = 5\%$), but with random feature weights.

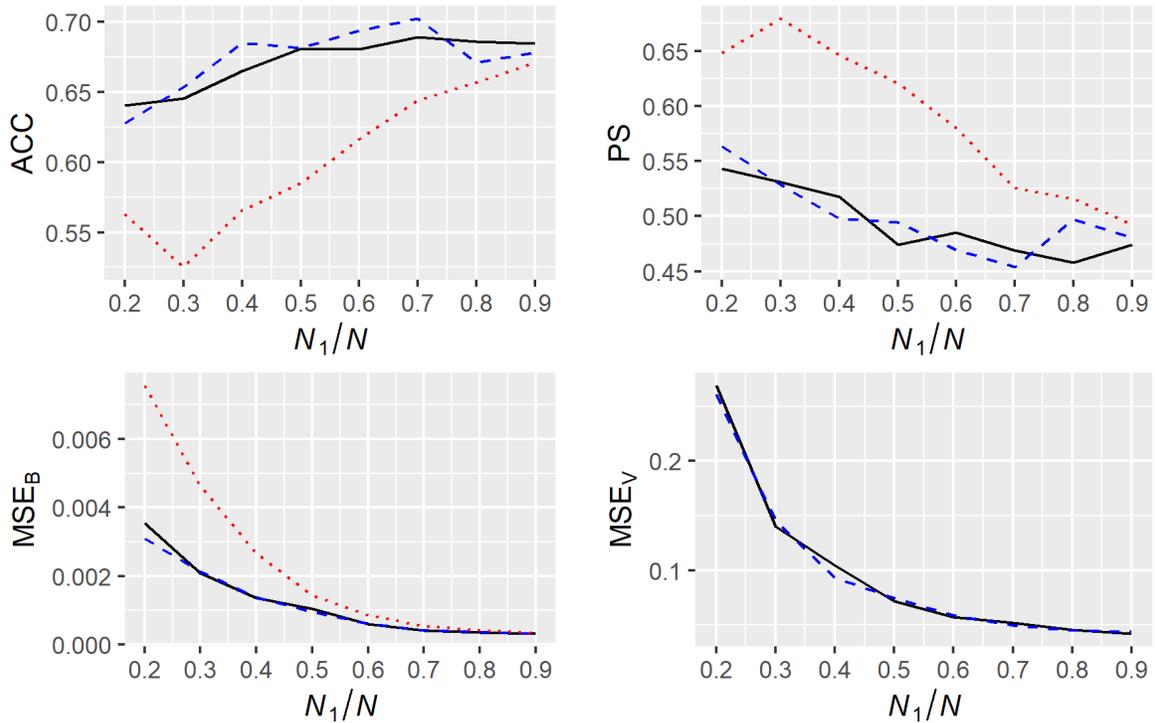

**Figure 9.** Similar to Figure 2 ($N = 100$, $\xi_1 = 40\%$, and $\xi_2 = 10\%$), but with random feature weights.

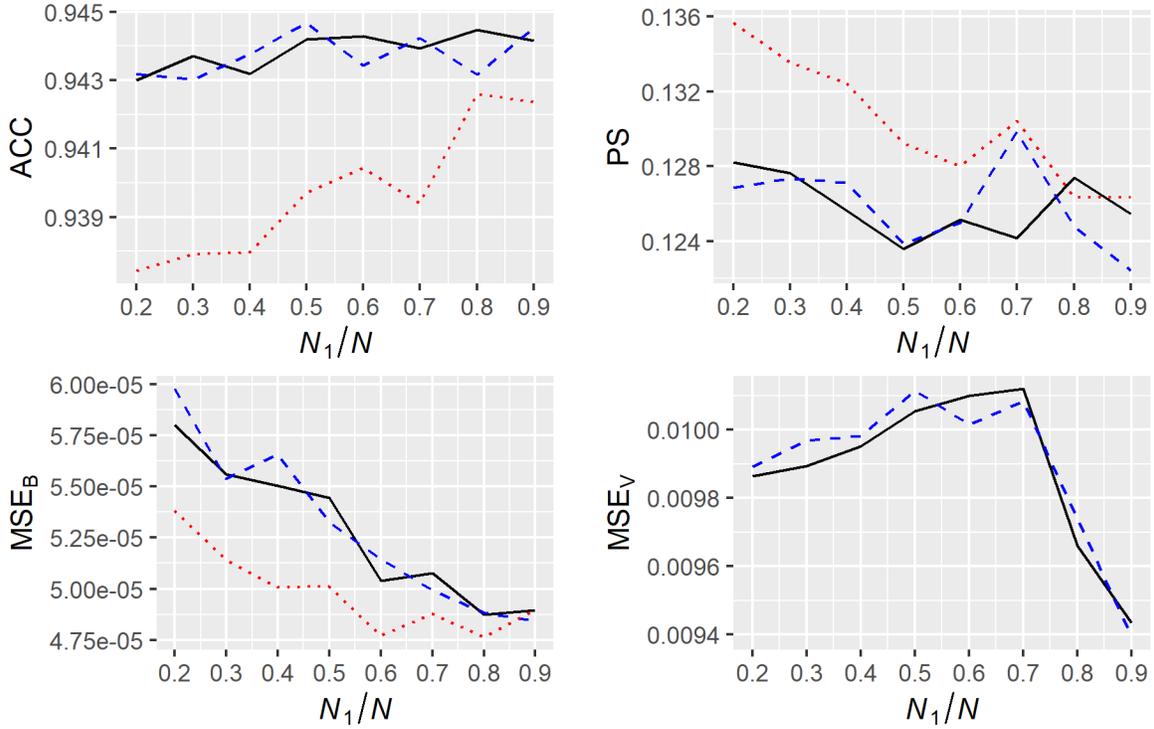

**Figure 10.** Similar to Figure 3 ($N = 500$, $\xi_1 = 40\%$, and $\xi_2 = 10\%$), but with random feature weights.

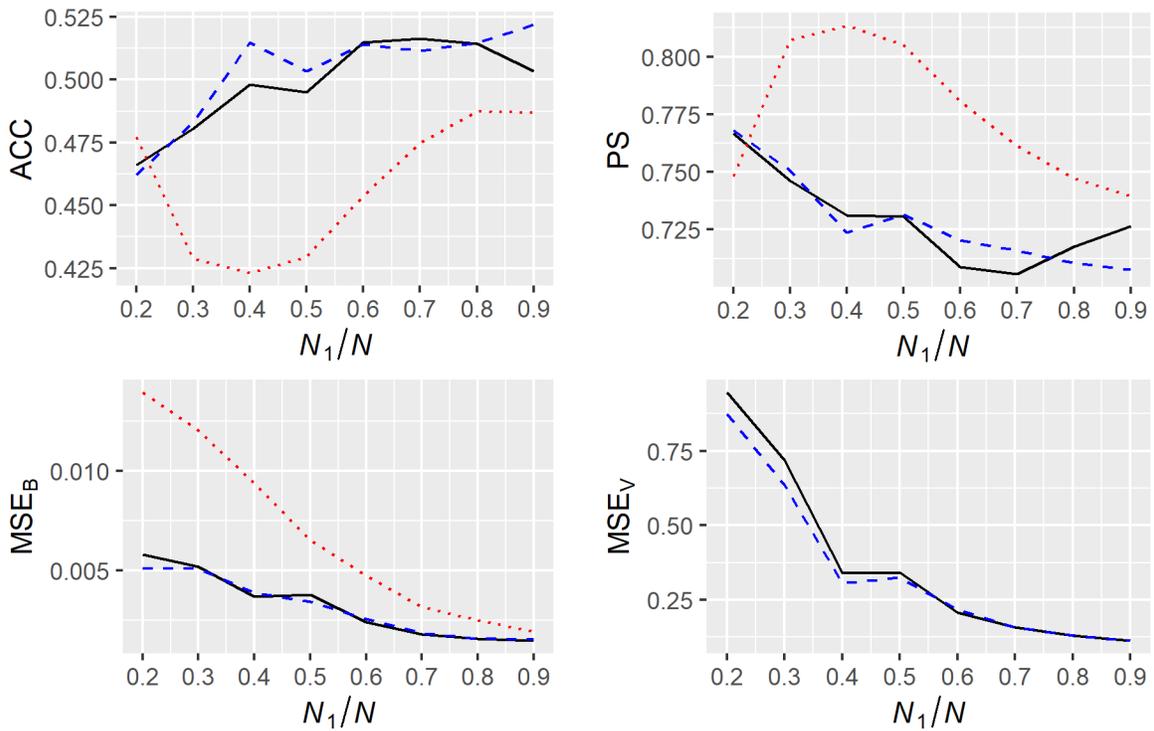

**Figure 11.** Similar to Figure 4 ($N = 100$, $\xi_1 = 60\%$, and $\xi_2 = 15\%$), but with random feature weights.

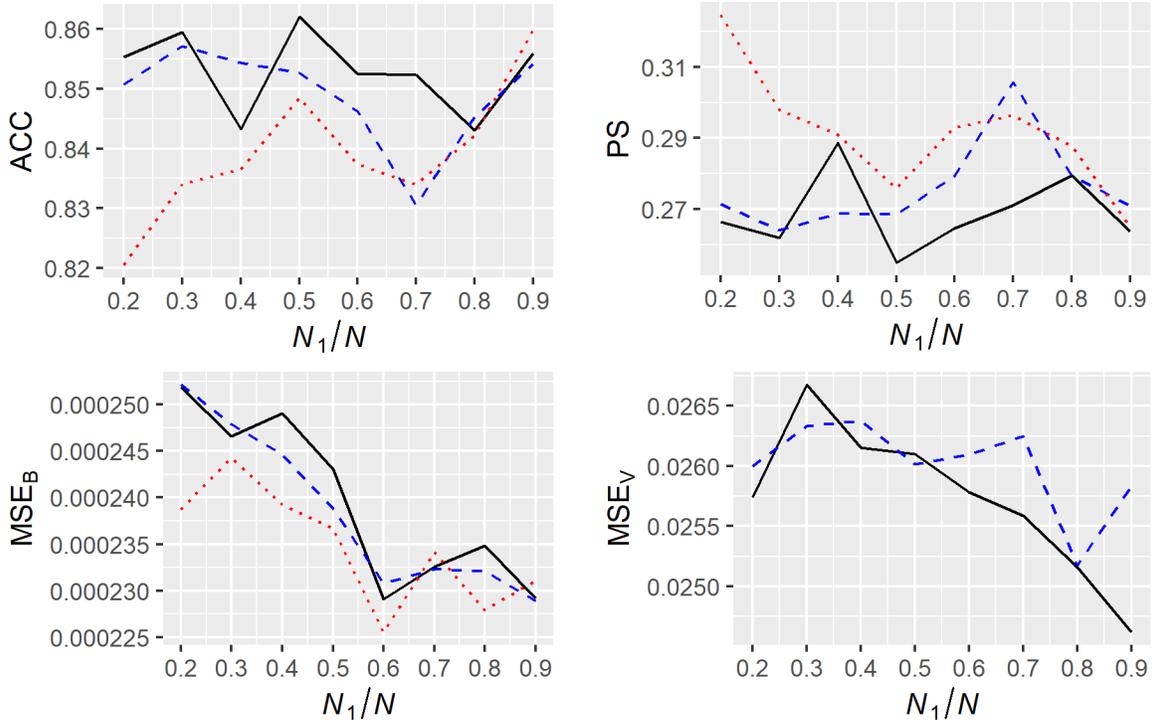

**Figure 12.** Similar to Figure 5 ($N = 500$, $\xi_1 = 60\%$, and $\xi_2 = 15\%$), but with random feature weights.

## G. Additional Experiments for the Kinship Term Example

In place of the gender variable, here we use the kinship degree variable as the known feature. For testing purposes, suppose that the kinship degree value for the cousin term is unknown. We use the method in Bui (2024) to initialize the proposed algorithm. Figure 4 plots the learned two-dimensional coordinates of the kinship terms. The gender feature can be inferred from Figure 4, as we can see quite clearly a group of male terms in the bottom left corner, a group of female terms in the top right corner, and the cousin term in the middle of these two groups. The generation gap feature can also be seen from Figure 4, in the direction from top-left to bottom-right of the plot. The imputed kinship degree value for Cousin is 3.91, which is close to 4 given in Table 2. These results also agree with those in Rosenberg and Kim (1975).

Now, we consider the generation gap variable as the known feature, with its value for Cousin treated as unknown for testing purposes. We again use the method in Bui (2024) to initialize the proposed algorithm. Figure 5 plots the learned two-dimensional coordinates of the kinship terms. The gender feature can be clearly recognized from Figure 5, as in Figure 4. The kinship degree feature can also be deducted in the direction from bottom-left to top-right corners of Figure 5. The imputed generation gap value for Cousin is approximately 0, agreeing with Table 2. These results again agree with those in Rosenberg and Kim (1975).

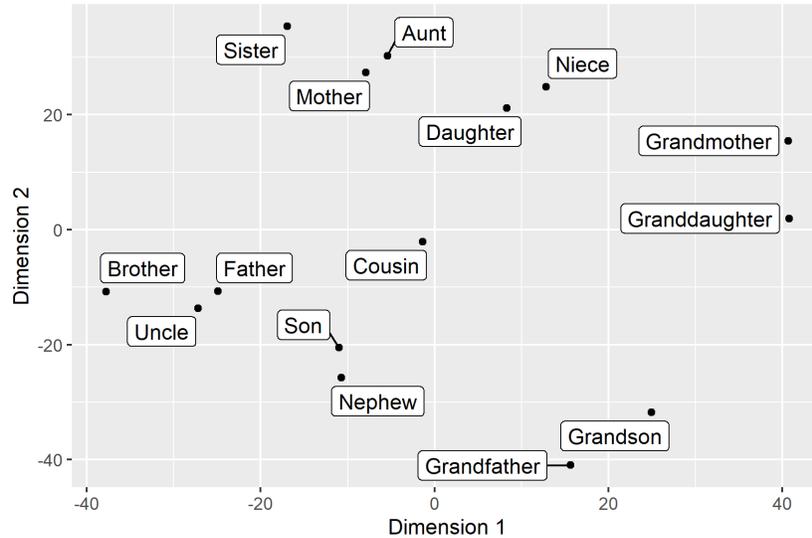

**Figure 13.** The two-dimensional coordinates of the 15 kinship terms learned by the proposed method, using Degree as the known feature (with its value for Cousin treated as unknown).

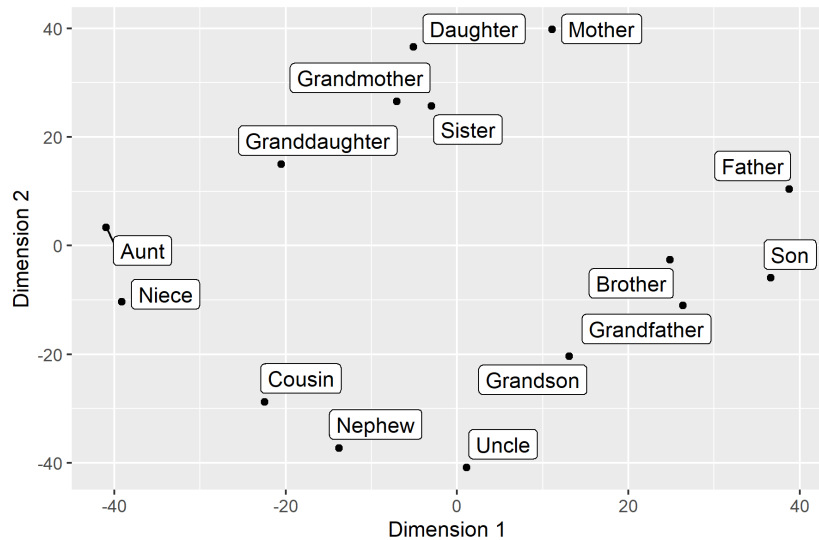

**Figure 14.** The two-dimensional **coordinates** of the 15 kinship terms learned by the proposed method, using generation gap as the known feature (with its value for Cousin treated as unknown).